%% file: paper.tex
\documentclass{article}
\usepackage{ijcai22}

\usepackage{times}
\usepackage{soul}
\usepackage{url}
\usepackage[hidelinks]{hyperref}
\usepackage[utf8]{inputenc}
\usepackage[small]{caption}
\usepackage{graphicx}
\usepackage{amsmath}
\usepackage{amsthm}
\usepackage{booktabs}
\usepackage{algorithm}
\usepackage{microtype}
\usepackage{algorithmicx}
\usepackage[noend]{algpseudocode}
\usepackage{amsfonts}
\usepackage{amssymb}
\usepackage{bbm}
\usepackage{bm}
\usepackage{color}
\usepackage{dirtytalk}
\usepackage{dsfont}
\usepackage{enumerate}
\usepackage{graphicx}
\usepackage{mathtools}
\usepackage{times}
\usepackage{wrapfig}
\usepackage{xspace}
\usepackage{booktabs}
\usepackage{accents}
\usepackage{caption}
\usepackage{subcaption}
\usepackage{float}
\usepackage{comment}

\usepackage[usenames,dvipsnames]{xcolor}
\hypersetup{
  pdffitwindow=true,
  pdfstartview={FitH},
  pdfnewwindow=true,
  colorlinks,
  linktocpage=true,
  linkcolor=Green,
  urlcolor=Green,
  citecolor=Green
}

\usepackage[capitalize,noabbrev]{cleveref}
\usepackage[backgroundcolor=white]{todonotes}
\newcommand{\commentout}[1]{}
\newcommand{\junk}[1]{}

\input{math_commands}

\title{\imo: Interactive Multi-Objective Off-Policy Optimization}

\author{
Nan Wang$^1$
\and
Hongning Wang$^1$\and
Maryam Karimzadehgan$^{2}$\and \\
Branislav Kveton$^3$\footnote{This work started prior to joining Amazon.}\and
Craig Boutilier$^2$
\affiliations
$^1$University of Virginia, 
$^2$Google Research,
$^3$Amazon
\emails
\{nw6a, hw5x\}@virginia.edu,
maryamk@google.com,
bkveton@amazon.com,
cboutilier@google.com
}

\begin{document}

\maketitle

\begin{abstract}
Most real-world optimization problems have multiple objectives. A system designer needs to find a policy that trades off these objectives to reach a desired operating point. This problem has been studied extensively in the setting of known objective functions. We consider a more practical but challenging setting of \emph{unknown objective functions}. In industry, this problem is mostly approached with online A/B testing, which is often costly and inefficient. As an alternative, we propose \emph{interactive multi-objective off-policy optimization} (\imo). The key idea in our approach is to interact with a system designer using policies evaluated in an \emph{off-policy fashion} to uncover which policy maximizes her unknown utility function. We theoretically show that \imo identifies a near-optimal policy with high probability, depending on the amount of feedback from the designer and training data for off-policy estimation. We demonstrate its effectiveness empirically on multiple multi-objective optimization problems.
\end{abstract}

\input{intro}

\input{problem}

\input{off-policy}

\input{algorithm}

\input{exp}

\input{related_work}

\input{conclusion}

\section*{Acknowledgments}
This work is partially supported by the National Science Foundation under grant IIS-2128009 and IIS-2007492, and by Google Research through the Student Researcher program. 

\bibliographystyle{named}
\bibliography{reference,elicitation_utility_models,long}

\appendix
\onecolumn
\input{appendix}

\end{document}

%% file: math_commands.tex
\Crefname{corollary}{Corollary}{Corollaries}
\Crefname{proposition}{Proposition}{Propositions}
\Crefname{theorem}{Theorem}{Theorems}
\Crefname{definition}{Definition}{Definitions}
\Crefname{assumption}{Assumption}{Assumptions}
\Crefname{example}{Example}{Examples}
\Crefname{remark}{Remark}{Remarks}
\Crefname{setting}{Setting}{Settings}
\Crefname{lemma}{Lemma}{Lemmas}
\Crefname{claim}{Claim}{Claims}

\usepackage{thmtools}
\declaretheorem[name=Theorem,refname={Theorem,Theorems},Refname={Theorem,Theorems}]{theorem}

\declaretheorem[name=Claim,refname={Claim,Claims},Refname={Claim,Claims}]{claim}

\newcommand{\cA}{\mathcal{A}}

\newcommand{\cW}{\mathcal{W}}
\newcommand{\cX}{\mathcal{X}}
\newcommand{\cV}{\mathcal{V}}

\newcommand{\realset}{\mathbb{R}}

\newcommand{\E}[1]{\mathbb{E} \left[#1\right]}

\newcommand{\prob}[1]{\mathbb{P} \left(#1\right)}

\newcommand{\normw}[2]{\|#1\|_{#2}}

\newcommand{\set}[1]{\left\{#1\right\}}

\newcommand{\T}{^\top}

\DeclareMathOperator*{\argmax}{arg\,max\,}
\DeclareMathOperator*{\argmin}{arg\,min\,}

\mathchardef\mhyphen="2D

\newcommand{\imo}{\ensuremath{\tt IMO^3}\xspace}

\newcommand{\ucb}{\ensuremath{\tt UCB1}\xspace}

%% file: intro.tex
\section{Introduction}
\label{sec:introduction}

Most real-world optimization problems involve multiple objectives. \emph{Multi-objective optimization (MOO)} has been studied and applied in various fields of system design, including engineering, economics, and logistics, where optimal policies need to trade off multiple, potentially conflicting objectives \cite{keeney-raiffa}. The system designer aims to find the optimal policy that respects her design principles, preferences and trade-offs. For example, when designing an investment portfolio, one's investment strategy requires trading off maximizing expected gain with minimizing risk \cite{liang2013portfolio}.

Two key issues need to be addressed in MOO before policy optimization. First, given a decision or \emph{policy space}, we need a mapping of policies to the expected values of the objectives in question. These objective values may be obtained by executing new policies on live traffic, which is risky and time-consuming \cite{deaton2018randomized,kohavi2009controlled}. A more efficient way for the mapping is through a model learned from data. In the example above, this might be a model specifying the expected return and the risk of an investment portfolio. In practice, acquiring data for learning a model can be costly and the model may be biased due to the data-gathering policy \cite{strehl2010learning}. Correcting for such biases is the target of the large literature on \emph{off-policy} statistical evaluation and optimization \cite{rosenbaum1983central,strehl2010learning,dudik2011doubly}. 
When objective values can be obtained for a new policy, \emph{bandit algorithms} can be used to optimize the policy  \cite{lattimore2020bandit}. However, these generally require scalar rewards that already dictate a decision maker's desired trade-offs among the different objectives.

This leads to the second issue---the specification of a single \emph{objective function} that dictates the desired trade-offs. In the example above, it might specify how much risk a decision maker can tolerate to attain some expected return. This can be viewed as the decision maker's \emph{utility function}. Assessing utility functions almost always requires interaction with the decision maker---requiring human judgements that typically cannot be learned from data in the usual sense \cite{keeney-raiffa}. Moreover, \emph{utility elicitation} is generally challenging and costly due to the cognitive difficulty faced by human decision makers when trying to assess trade-offs among objectives in a quantitatively precise fashion \cite{tver-kahn:1974,camerer:2003}. While some elicitation techniques attempt to identify the full objective function \cite{keeney-raiffa}, others try to minimize this burden in various ways. One common principle is to limit trade-off assessments to only those that are relevant given the \emph{feasible or realizable} combinations of objectives w.r.t.\ the utility model and policy constraints \cite{boutilier:regretSurvey2013}.\footnote{Much work in MOO focuses on the identification of \emph{Pareto optimal} solutions---those that induce a vector of objective values such that no single objective can be improved without degrading another \cite{mascolell:book}. The selection of a solution from this set still requires the decision maker to choose and thus make a trade-off, possibly implicitly.} This requires that the model is known.

We propose an interactive off-policy technique that supports a system designer in identifying the optimal policy that trades off multiple objectives with an \emph{unknown} utility function \cite{branke2008multiobjective}. The key is to address the dependence between querying the utility function to more effectively learn a model, and using a model to more efficiently elicit the utility function. The utility function is modeled as a linear scalarization of multiple objectives \cite{keeney-raiffa}, where the scalarization parameters specify the trade-off among the objectives. We use off-policy estimators to evaluate policies in an unbiased way \emph{without ever executing them}. To learn the model, we present the off-policy estimates of the objective values of candidate policies to the designer for feedback. The candidate policies are chosen judiciously to maximize the information gain from the feedback. 
Over time, (i) the model converges to the trade-offs embodied in the unknown utility function by learning from the designer's feedback; and (ii) the policy induced by the model converges to the optimal policy. We analyze our approach and prove theoretical guarantees for finding the near-optimal policies. Our comprehensive empirical evaluation on four multi-objective optimization problems shows the effectiveness of our method.


%% file: problem.tex
\section{Problem Formulation}
\label{sec:problem}
For simplicity, we use $[n]$ to denote the set $\{1, \dots ,n\}$. Consider a policy optimization problem with $d \geq 1$ \emph{(potentially conflicting) objectives}. Let $\cX$ be a \emph{context space} and $\cA$ an \emph{action space} with $K$ actions. In each round, $x \in \cX$ is sampled from a context distribution $P_x$. 
An action $a\in\cA$ is taken in response following a \emph{stochastic policy} $\pi(\cdot \mid x)$, which is a distribution over $\cA$ 
for any $x\in\cX$. The policy space is $\Pi = \big\{\pi \,\big|\, \pi(\cdot \mid x) \in \Delta_{K - 1}, \forall x\in\cX \big\}$, where $\Delta_{K}$ is the $K$-simplex with $K + 1$ vertices. After taking action $a$, the agent receives a $d$-dimensional reward vector $r \in [0,1]^d$ sampled from a reward distribution $P_{r}(\cdot \mid x,a)$, corresponding to $d$ objectives. The \emph{expected value} of policy $\pi$ is $V(\pi)=\mathbb E_{x \sim P_x, a\sim\pi(\cdot|x), r \sim P_r(\cdot \mid x,a)}\big[r\big]$. Note that $V(\pi)$ is a $d$-dimensional vector whose $i$-th entry $V_i(\pi)$ is the expected value of objective $i$ under policy $\pi$.

We assume that there exists a \emph{utility function} $u_{\theta}$, parameterized by $\theta$, which is used by the designer to assess the quality $u_{\theta}(v)$ of any objective-value vector $v \in \realset^d$. Without loss of generality, $u_{\theta}$ is \emph{absolutely monotonic} in each objective; but the correlations and conflicts among the objectives are unknown. We adopt the common assumption that $u_\theta$ is linear \cite{keeney-raiffa} and determined by a \emph{scalarization} $u_{\theta}(v) = \theta\T v$ of the objective values, where $\theta \in \realset^d$ determines the designer's trade-off among the objectives. We treat $\theta$ as \emph{a priori} unknown, and moreover that it cannot be (easily) specified directly by the designer. Hence, we learn it through the \emph{interactions} with the designer.

The \emph{optimal policy}, for any fixed designer's trade-off preferences $\theta_* \in \realset^d$, is defined as
\begin{equation}
  \pi_*
  = \argmax_{\pi \in \Pi} u_{\theta_*}\big(V(\pi)\big)\,.
  \label{eq:optimization}
\end{equation}
Since the interactions can be costly, we consider a fixed budget of $T$ rounds of interactions with the designer. Our goal is to find a near-optimal policy with high probability after the interactions. Specifically, we use \emph{simple regret} \cite{lattimore2020bandit} to measure the optimality of a policy $\pi$, which is the difference in the utilities of $\pi_*$ and $\pi$,
\begin{equation}
  R_T^{sim}
  = u_{\theta_*}\big(V(\pi_*)\big) - u_{\theta_*}\big(V(\pi)\big)\,.
  \label{eq:simpleregret}
\end{equation} 
Here we only focus on the quality of the best policy identified after these interactions, not the quality of policies presented during interactions.


\section{General Algorithm Design}
\label{sec:general design}

We first describe our approach in general terms, motivating it by the \emph{de facto} standard approach to A/B testing in industry \cite{kohavi2009controlled}. In the standard \say{iterative} approach, a policy designer proposes a candidate policy $\pi$ and evaluates it on live traffic for some time period (say, two weeks, to average out basic seasonal trends). If $\pi$ outperforms a \emph{production} policy (e.g., it improves some metrics/objectives and does not degrade others, or it achieves a desired trade-off among all objectives), $\pi$ is \emph{accepted} and deployed. If it does not, the designer proposes a new candidate policy and the process is repeated. This approach has three major shortcomings. First, each iteration takes a long time and many iterations may be needed to find a good policy. Second, it is difficult to propose good candidate policies, because the policy space is large and it is not a priori clear which objective trade-offs are feasible. Finally, due to the difficulty of managing changes in the control and treatment groups in large-scale platforms, online randomized experiments often lead to unexpected results \cite{kohavi2009controlled,kohavi2011unexpected}, which limit its efficiency and application in the fast-evolving industrial settings.

Now consider an idealized scenario where the designer knows $V(\pi)$ for any policy $\pi \in \Pi$. Then we could learn $\theta_*$ in \eqref{eq:optimization} by interacting with the designer. A variety of preference elicitation techniques could be used \cite{keeney-raiffa,boutilier:regretSurvey2013}. We study the following approach. In round (interaction) $t$, we (i) propose a policy $\pi_t$; (ii) present the value vector $V(\pi_t)$ to the designer; and (iii) obtain a noisy response based on the designer's true utility $u_{\theta_*}(V(\pi_t))$.
The feedback can take different forms, but ultimately reflects the designer's perceived value for $\pi_t$. We assume a binary feedback of the form \say{Is policy $\pi$ \emph{acceptable}?}, motivated by our industry example above.

In this work, we consider a more realistic but also more challenging setting where $V(\pi)$ is unknown. In principle, any policy $\pi$ can be evaluated on live traffic. However, online evaluation can be costly, inefficient, and time consuming; leading to unacceptable delays in finding $\pi_*$ \cite{deaton2018randomized,kohavi2009controlled}. To address this issue, we evaluate $\pi$ \emph{offline} using logged data generated by some prior policy, such as the production policy \cite{adith2015counterfactual}. In \cref{sec:off-policy}, we introduce three most common off-policy estimators for this purpose. The off-policy estimated value vector $\hat V(\pi_t)$ is then used in the elicitation process with the designer. Finally, we learn $\theta_*$ and $\pi_*$ based on the estimated values and \emph{noisy feedback} from interactions with the designer. We present our algorithm and analyze it in \cref{sec:algorithm}.

%% file: off-policy.tex
\section{Multi-Objective Off-Policy Evaluation and Optimization}
\label{sec:off-policy}

In this section, we discuss how to evaluate a policy $\pi$ using logged data generated by another (say, production) policy, and optimize $\pi$ w.r.t.\ any (fixed and known) scalarization parameters $\theta$. We have a set of logged records $\mathcal{D} = \big\{(x_j, a_j, r_j)\big\}_{j = 1}^N$ collected by a \emph{logging policy} $\pi_0$ as an input. For the $j$-th record, $x_j$ is the context, $a_j$ is the action from $\pi_0$, and $r_j$ is the realized reward vector. We also assume that \emph{propensity scores} $\pi_0(a_j \mid x_j)$ (i.e., the probability that $\pi_0$ takes action $a_j$ given context $x_j$) are logged. If not, they can be estimated from logged data \cite{strehl2010learning}.

\subsection{Evaluation}
\label{sec:evaluation}
Off-policy evaluation has been studied extensively in the single-objective setting \cite{strehl2010learning,dudik2011doubly}. Generally, better evaluation leads to better optimization \cite{strehl2010learning}. By treating the reward as a $d$-dimensional vector rather than a scalar, we can adapt existing off-policy estimators to MOO. We adapt 
three
popular estimators below.

The first estimator, the \emph{direct method (DM)} \cite{diane2007more}, estimates the expected reward vector $\mathbb E\big[r \mid x,a \big]$ by $\hat r(a, x) \in \realset^d$, where $\hat{r}$ is some offline-learned reward model. The policy value is estimated by 
\begin{align}
\label{eq:DM}
  \hat V^\textsc{dm}(\pi)
  = \frac{1}{N} \sum_{j = 1}^N \sum_{a\in \cA} \pi(a \mid x_j) \hat r(a, x_j)\,.
\end{align}
Since the model is learned without knowledge of $\pi$, it may focus on areas that are irrelevant for $V(\pi)$, resulting in a biased estimate of $V(\pi)$ \cite{beygelzimer2016offset}.

The second estimator, \emph{inverse propensity scoring (IPS)} \cite{rosenbaum1983central}, is less prone to bias. Instead of estimating rewards, IPS uses the propensities of logged records to correct the shift between the logging and new policies, 
\begin{align}
\label{eq:IPS}
  \hat{V}^\textsc{ips}(\pi)
  = \frac{1}{N} \sum_{j = 1}^N \min \left\{M,
  \frac{\pi(a_j \mid x_j)}{\pi_0(a_j \mid x_j)}\right\} r_j\,,
\end{align}
where $M > 0$ is a hyper-parameter that trades off the bias and variance in the estimate. The IPS estimator is unbiased for $M = \infty$, but can have a high variance if $\pi$ takes actions that are unlikely under $\pi_0$. When $M$ is small, the variance is small but the bias can be high, since the IPS scores are clipped.


To alleviate the high variance of IPS, we can take advantage of both $\hat r$ and IPS to construct the \emph{doubly robust (DR)} estimator \cite{dudik2011doubly}
\begin{align}
  \label{eq:DR}
  \hat V^\textsc{dr}(\pi) \!=\!
  \frac{1}{N}\!\! \sum_{j = 1}^N \frac{\pi(a_j \!\mid\! x_j)}{\pi_0(a_j \!\mid\! x_j)}
  (r_j \! - \! \hat r(a_j,\! x_j))\! +\! \hat V^\textsc{dm}(\pi).
\end{align} 
Intuitively, $\hat r$ is used as a baseline for the IPS estimator. If the model for reward estimation is unbiased or the propensities are correctly specified, DR can provide an unbiased estimate of the value. It has been shown that DR achieves lower variance than IPS \cite{dudik2011doubly}.

\subsection{Optimization}
\label{sec:optimization}

A key component in our approach is \emph{policy optimization}, i.e., finding the optimal policy given a scalarization vector $\theta$,
\begin{equation}
  \hat{\pi}
  = \argmax_{\pi \in \Pi} u_\theta\big(\hat{V}(\pi)\big) = \argmax_{\pi \in \Pi} \theta\T \hat{V}(\pi)\,,
  \label{eq:off-optimization}
\end{equation} 
where $\hat V(\pi)$ is some off-policy estimator. The optimized variables are the entries of $\pi \in \Pi$ that represent the probabilities of taking actions. In \cref{appendix:linearprog}, we prove that \eqref{eq:off-optimization} can be formulated as a \emph{linear program (LP)} for all off-policy estimators in \cref{sec:evaluation} in the tabular case, where the policy is parameterized separately for each context. For non-tabular policies, we suggest using gradient-based policy optimization methods \cite{adith2015counterfactual}, though we provide no theoretical guarantees for this case.

Since \eqref{eq:off-optimization} is an LP for all our estimators, at least one solution to \eqref{eq:off-optimization} is a vertex of the feasible set,
corresponding to 
\emph{non-dominated policies}, which cannot be written as a convex combination of other policies. For such policies, we can ``learn'' $\pi_*$ by first learning $\theta_*$ then optimizing the policy under $\theta_*$. We now turn to the question of estimating $\theta_*$ using interactive designer feedback.

%% file: algorithm.tex
\section{Interactive Multi-Objective Off-Policy Optimization}
\label{sec:algorithm}
Off-policy estimation and optimization in \cref{sec:off-policy} assume that the parameters $\theta_*$ are known. Now we turn to \emph{interactively} estimating $\theta_*$ by querying the designer for feedback on carefully selected policies over $T$ rounds. Utility elicitation can be accomplished using a variety of query formats (e.g., value queries, bound queries, $k$-wise comparisons, critiques) and optimization criteria for selecting queries \cite{keeney-raiffa,preference:aaai02,boutilier:regretSurvey2013}. 

\subsection{Query Model}
Following a common industrial practice (\cref{sec:general design}), we adopt a simple query model where we ask the designer to rate an objective value vector $v$ corresponding to $d$ objectives as \say{\emph{acceptable}} or \say{\emph{not acceptable}.} We require a \emph{response model} that relates this stochastic feedback to the designer's underlying utility for $v$. We adopt a \emph{logistic response} model
\begin{align}
  \ell_{\theta_*}(v)
  = 1 / (1+\exp(-u_{\theta_*}(v)))\,,
  \label{eq:query model}
\end{align}
where $u_{\theta_*}(v) = \theta_*\T v$, and the designer responds \say{\emph{acceptable}} with probability $\ell_{\theta_*}(v)$ and \say{\emph{not acceptable}} otherwise. 
Roughly speaking, this can be understood as a designer's noisy feedback relative to some implicit baseline (e.g., the value vector of the production policy). 
Logistic response of this form arises frequently in modeling binary or $k$-wise discrete choice in econometrics, psychometrics, marketing, AI, and other fields \cite{mcfadden_condlogit:1974,viappiani:nips2010}; and lies at the heart of feedback mechanisms in much of the dueling bandits literature \cite{miroslav2015contextual}. We defer the study of other types of feedback to future work.

\subsection{\imo: Interactive Multi-objective Off-Policy Optimization}
Now we introduce \imo for engaging the designer in solving the MOO problem. We approach the problem as fixed-budget \emph{best-arm identification (BAI)} \cite{karnin2013almost}, where we minimize the simple regret \eqref{eq:simpleregret} in $T$ rounds of interaction. At a high level, \imo works as follows. In round $t \in [T]$, it selects a policy (arm) $\pi_t$ and presents its off-policy estimated value vector $\hat V(\pi_t)$ to the designer. The designer responds with $Y_t \sim \mathrm{Ber}\big(\ell_{\theta_*}(\hat V(\pi_t))\big)$. After $T$ rounds, we compute the maximum likelihood estimate (MLE) $\hat\theta$ of $\theta_*$,  where $\hat V(\pi_t)$ serves as a feature vector for response $Y_t$.

To make \imo statistically efficient in identifying the optimal policy with limited budget, we must design a good distribution over policies to be presented to the designer. One challenge is that the policy space $\Pi$ is continuous and infinite. To address this issue, we first discretize $\Pi$ to a set $\cW$ of $L$ diverse policies, which are optimal under different random scalarizations. The other challenge is learning $\theta_*$ efficiently. We approach this as an optimal design problem \cite{wong1994comparing}. Specifically, we use \emph{G-optimality} to design a distribution over $\cW$, from which we draw $\pi_t$ in round $t$ that minimizes variance of the MLE $\hat\theta$. Since the design is variance minimizing, \imo chooses the final optimal policy $\tilde\pi_*$ solely based on the highest mean utility under $\hat\theta$. We experimented with more complex algorithm designs, where the distribution of $\pi_t$ was adapted with $t$, analogous to sequential halving in BAI \cite{karnin2013almost,arthur2016non-stochastic}. However, none of these approaches improved \imo, and
thus we focus on the non-adaptive algorithm. 

We present \imo in \cref{alg:imo3}. In lines 1--5, the policy space $\Pi$ is discretized into $L$ policies $\cW$. Each policy in $\cW$ is optimal under some $\theta_i$. Since $\theta_i$ are sampled uniformly from a unit ball, representing all scalarization directions, the policies $\pi_i$ are diverse and allow us to learn about any $\theta_*$ efficiently. In our regret analysis, we assume that $\hat{\pi}_*$ in \eqref{eq:off-optimization} under $\theta_*$ is in $\cW$. Note that we do not interact with the designer in this stage. In line 6, we compute the \emph{G-optimal design} over $\cW$, a distribution over $\cW$ that minimizes the variance of the MLE $\hat\theta$ after $T$ rounds. In lines 7--10, we interact with the designer over $T$ rounds. In round $t \in [T]$, we draw $\pi_t$ according to the G-optimal design, present its values $\hat V(\pi_t)$ to the designer, and receive feedback $Y_t$. In line 11, we compute the MLE $\hat{\theta}$ from all collected observations $\{\hat V(\pi_t), Y_t\}_{t = 1}^T$. Finally, we use the estimated $\hat\theta$ to find the identified optimal policy $\tilde\pi_*$ w.r.t.\ off-policy estimated values using an LP (\cref{sec:optimization}).

\begin{algorithm}
\caption{\imo}
\label{alg:imo3}
\begin{flushleft}
\hspace*{\algorithmicindent}\textbf{Input:} Logging policy $\pi_0$, logged data $\mathcal{D}$, budget $T$, \\ 
\hspace*{\algorithmicindent} \qquad\;\; and pre-selection budget $L$
\end{flushleft}
\begin{algorithmic}[1]
\State $\mathcal W \gets \{\}$
\For{$i=1,\dots,L$} 
    \State Sample $\theta_i$ from a unit ball in $\realset^d$
    \State $\pi_i \gets \argmax_{\pi\in\Pi} u_{\theta_i} \big(\hat V(\pi)\big)$ 
    \State $\mathcal W \gets \mathcal W + \{\pi_i\}$
\EndFor
\State $P_G(\mathcal W) \gets$ G-optimal design over $\mathcal W$
\For{$t=1,\dots,T$} 
    \State $\pi_t \sim P_G(\mathcal W)$
    \State Present $\hat V(\pi_t)$ to the designer and observe $Y_t$ 
    \EndFor
\State $\hat\theta \gets \mathrm{MLE}(\{\hat V(\pi_t), Y_t\}_{t=1}^{T})$
\State Return $\tilde\pi_* \gets \argmax_{\pi\in\Pi} u_{\hat\theta} \big(\hat V(\pi)\big)$
\end{algorithmic}
\end{algorithm}

\subsection{Regret Analysis}

We now analyze the simple regret of \imo, which is defined in \eqref{eq:simpleregret}. Due to space constraints, we focus on the IPS estimator and then discuss extensions to other estimators.

To state our regret bound, we first introduce some notations. $\cW = \set{\pi_i}_{i = 1}^L$ is the pre-selected policies in \imo and $\cV = \set{v_i}_{i = 1}^L$ is their estimated values, with $v_i = \hat{V}(\pi_i)$. Let the optimal policy $\hat{\pi}_*$ under $\hat{V}(\pi)$ be in $\cW$ and $\pi_1 = \hat{\pi}_*$ without loss of generality. Let $\mu_i = v_i\T \theta_* \in [0, 1]$ be the utility of policy $\pi_i$ and $\Delta_i = \mu_1 - \mu_i$ be its gap. $\Delta_{\min} = \min_{i > 1} \Delta_i$ denotes the minimum gap. Let $\alpha_* = \argmin_{\alpha \in \Delta_{L - 1}} g(\alpha)$ be the \emph{G-optimal design} on $\cV$, where $g(\alpha) = \max_{i \in [L]} v_i\T G_\alpha^{-1} v_i$ and $G_\alpha = \sum_{i = 1}^L \alpha_i v_i v_i\T$. Let $h(\cdot)$ be the sigmoid function and $h'(\cdot)$ be its derivative. 

\begin{restatable}[]{theorem}{regretanalysis}
\label{theorem:regretanalysis} 
Let $c_{\min}, \delta_1 > 0$ be chosen such that
\begin{align*}
  \min_{v \in \cV} \min \{h'(v\T \theta_*), h'(v\T \hat{\theta})\}
  \geq c_{\min}
\end{align*}
holds with probability at least $1 - \delta_1$. Then $R_T^{sim} \leq$
\begin{align}
  L \exp\left[- \frac{\Delta_{\min}^2 c_{\min}^2 T}{2 g(\alpha_*)}\right] +
  2 ||\theta_*||_2 \sqrt{\frac{d M^2 \log(2 d /\delta_2)}{2 N}}
  \label{eq:regret bound}
\end{align}
holds with probability at least $1 - (\delta_1 + 2\delta_2)$, where $d$ is the number of objectives, $M$ is the tunable parameter in the IPS estimator, and $N$ is the size of logged data.
\end{restatable}

The proof of \cref{theorem:regretanalysis} is in \cref{appendix:proof}. The regret bound decomposes into two terms. The first term is the regret of BAI w.r.t.\ \emph{estimated policy values} and decreases with the amount of designer's feedback $T$. The second term reflects the error of the IPS estimator and decreases with data size $N$.

Specifically, the first term in \eqref{eq:regret bound} is $O(L \exp[- T])$. While it increases with the number of pre-selected policies $L$, it decreases exponentially with budget $T$. Therefore, even relatively small sample sizes of $T = O(\log L)$ lead to low simple regret. Since we assume that the optimal policy $\hat{\pi}_*$ under $\hat{V}(\pi)$ is in $\cW$, $L$ needs to be large for this condition to hold. Regarding the other terms, $\Delta_{\min}^2 c_{\min}^2$ is a problem-specific constant and we minimize $g(\alpha_*)$ by design.

The second term in \eqref{eq:regret bound} decreases with data size $N$ at an expected rate $O(\sqrt{1 / N})$. Now we discuss the errors for other estimators. For the DM estimator, this error depends on the quality of the model and can not be directly analyzed. It could be large when the model is biased. For the DR estimator, it is unbiased if the reward model is unbiased or the propensity scores are correctly specified. If the model is unbiased, there is no error in the DR estimator. Otherwise, the error is bounded 
as in the IPS estimator.

%% file: exp.tex
\section{Experiments}
\label{sec:exp}
In this section, we evaluate \imo on four MOO problems. We introduce the problems for evaluation in \cref{sec:moo-probs}, describe several baseline methods in \cref{sec:baselines}, and evaluate \imo vs.\ baselines from different perspectives in \cref{sec:results}. All the datasets and implementations used in experiments will be made public upon publication of the paper.

Due to space limit, we put the details of how to generate logged data for each problem in \cref{appendix:exp}. To simulate designer feedback, we sample the ground-truth scalarization $\theta_*\in\realset^2$ from the unit ball, and sample responses from $\mathrm{Ber}\big(\ell(\hat V(\pi);\theta_*)\big)$, where $\hat V(\pi)$ is off-policy estimated value vector presented to the designer. We generate feedback in the same way in all four problems.  
\begin{figure*}
    \centering
    \begin{subfigure}[b]{0.245\textwidth}
        \centering
        \includegraphics[width=\linewidth]{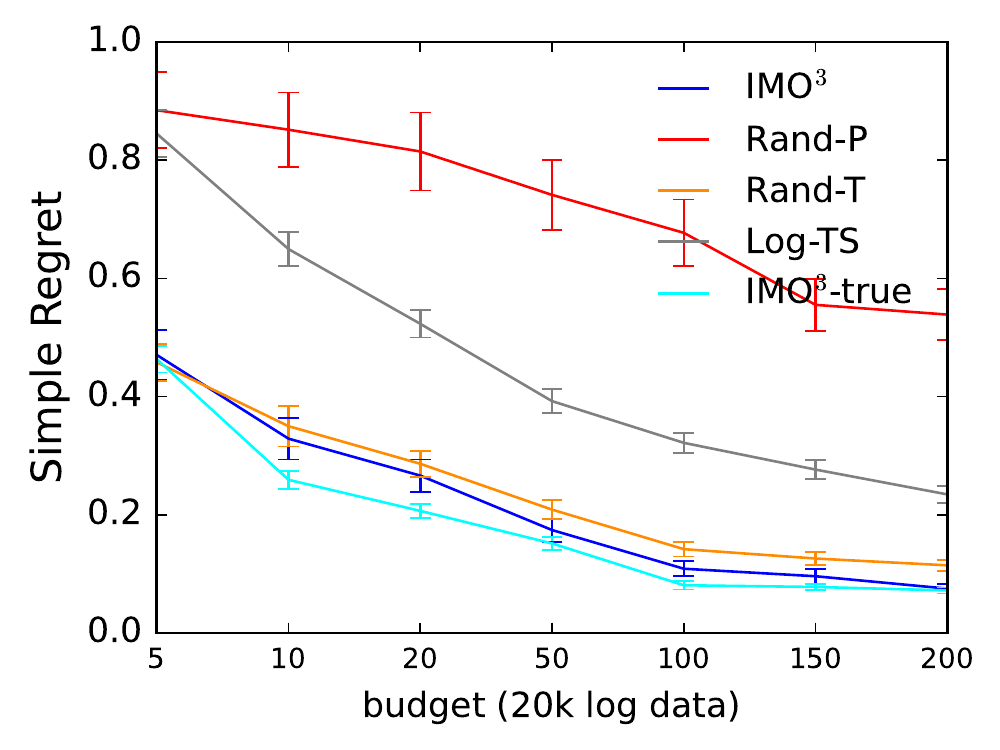}
        \caption{ZDT1. }
        \label{fig:zdt1-varybudget}
    \end{subfigure}
    \begin{subfigure}[b]{0.245\textwidth}
        \centering
        \includegraphics[width=\linewidth]{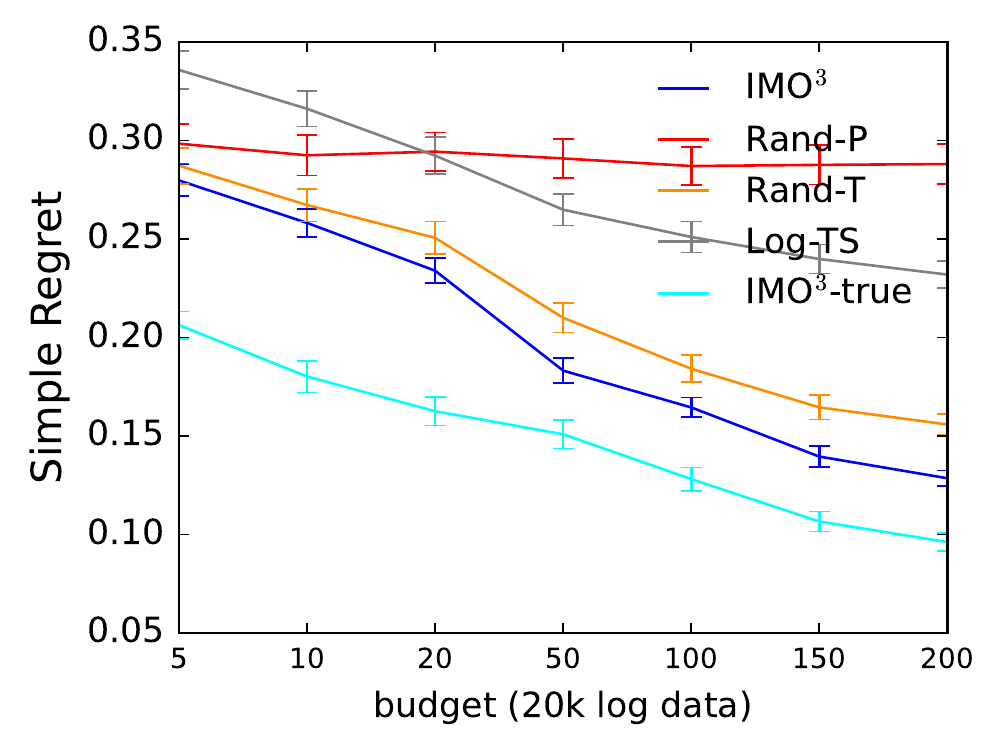}
        \caption{Crashworthiness.}
        \label{fig:crash-varybudget}
    \end{subfigure}
    \begin{subfigure}[b]{0.245\textwidth}
        \centering
        \includegraphics[width=\linewidth]{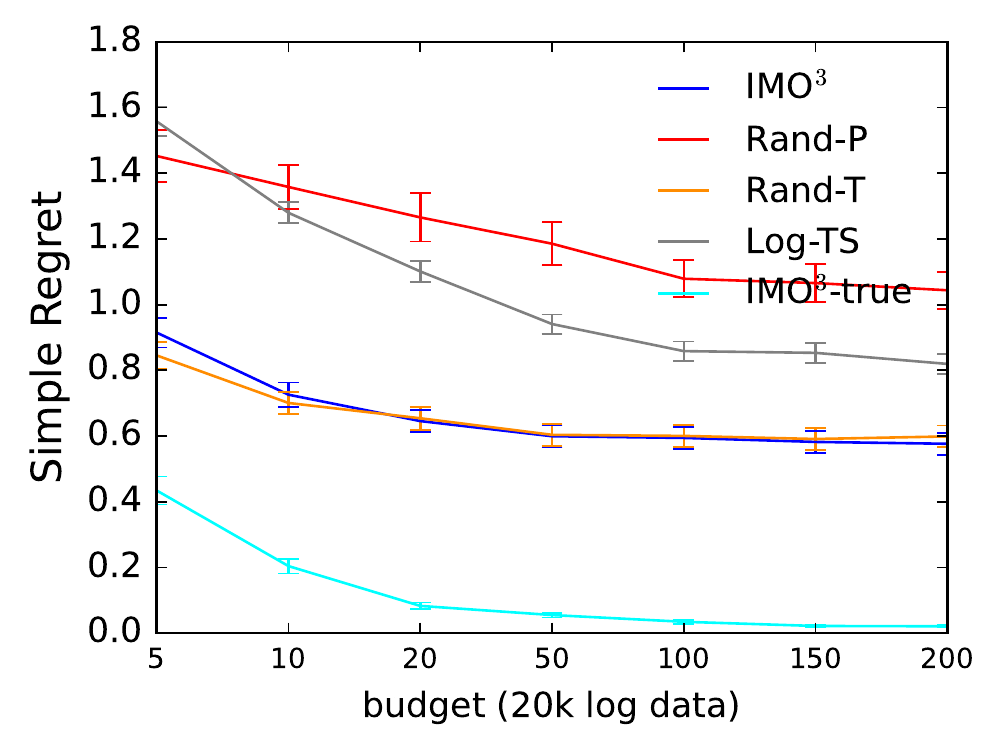}
        \caption{Stock investment. }
        \label{fig:stock-varybudget}
    \end{subfigure}
    \begin{subfigure}[b]{0.245\textwidth}
        \centering
        \includegraphics[width=\linewidth]{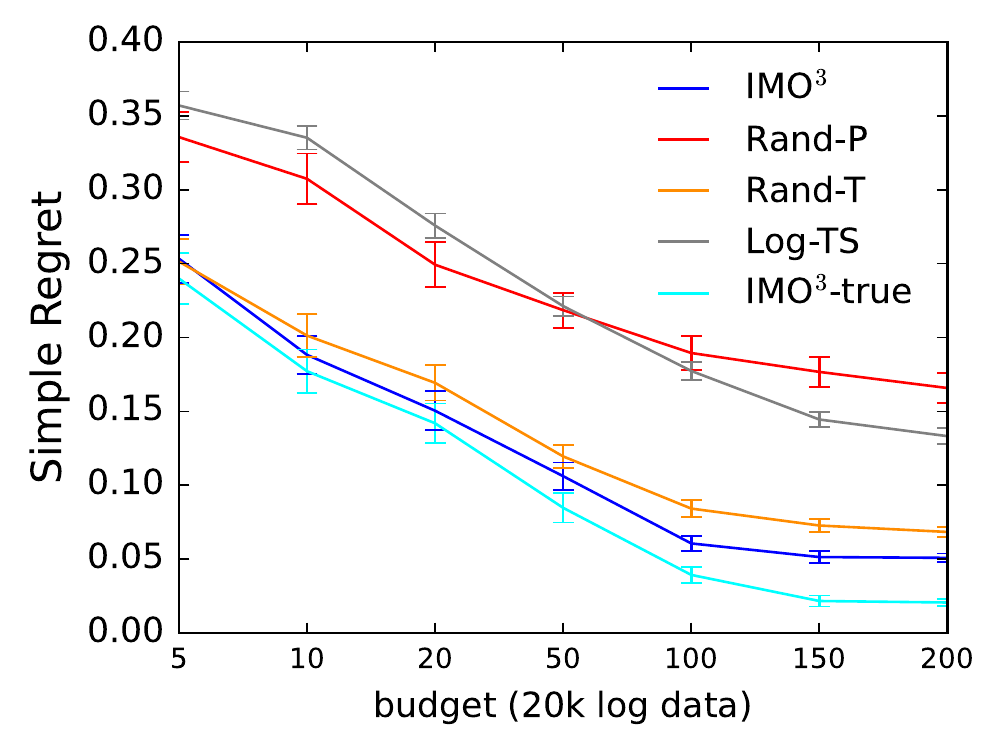}
        \caption{Yahoo! news recommendation.}
        \label{fig:click-varybudget}
    \end{subfigure}
    \caption{Simple regret of different algorithms by fixing logged data size $N=20,000$ and varying budget. Each experiment is averaged over 10 logged data, 10 randomly selected $\theta_*$ and 5 runs under each combination of logged data and $\theta_*$.}
\label{fig:varybudget}
\end{figure*}

\subsection{Multi-Objective Optimization Problems}
\label{sec:moo-probs}
\noindent
\textbf{ZDT1.}\hspace*{1mm}
The ZDT test suite \cite{zitzler2000comparison} is the most widely employed benchmark for MOO. We use ZDT1, the first problem in the test suite,
a box-constrained $n$-dimensional two-objective problem, with objectives $F_1$ and $F_2$ defined as
\begin{align}
    & F_1(x) = 5x_1, \quad F_2(x) = g(x) \bigg[1-\sqrt{\frac{x_1}{g(x)}}\bigg]\,, \\\nonumber
    & \textrm{and }\; g(x) = 1 + \frac{9(\sum_{i=2}^n x_i)}{n-1}\,,
\end{align}
where $x=(x_i)_{i=1}^n$ are variables and $x_i\in [0,1], \forall i\in[n]$.
We use $n=5$ in our experiments, treating $(x_4, x_5)$ as context, and perform optimization on $(x_i)_{i=1}^3$. We sample five combinations of $(x_4, x_5)$ uniformly to create context set $\cX$ and ten combinations of $(x_i)_{i=1}^3$ to create the action set $\cA$. 
\\\noindent
\textbf{Crashworthiness.}\hspace*{1mm}
This MOO problem is extracted from a real-world crashworthiness domain \cite{carvalho2018solvingRM}, where three objectives factor into the optimization of the crash-safety level of a vehicle. 
We refer 
to Sec.~2.1 of \cite{carvalho2018solvingRM} for detailed objective functions and constraints. 
Five bounded decision variables $(x_i)_{i=1}^5$  represent the thickness of reinforced members around the car front. We  use different combinations of the last two variables as contexts and the first three 
as actions. The rest settings are the same as for ZDT1. 
\begin{figure*}
    \centering
    \begin{subfigure}[b]{0.245\textwidth}
        \centering
        \includegraphics[width=\linewidth]{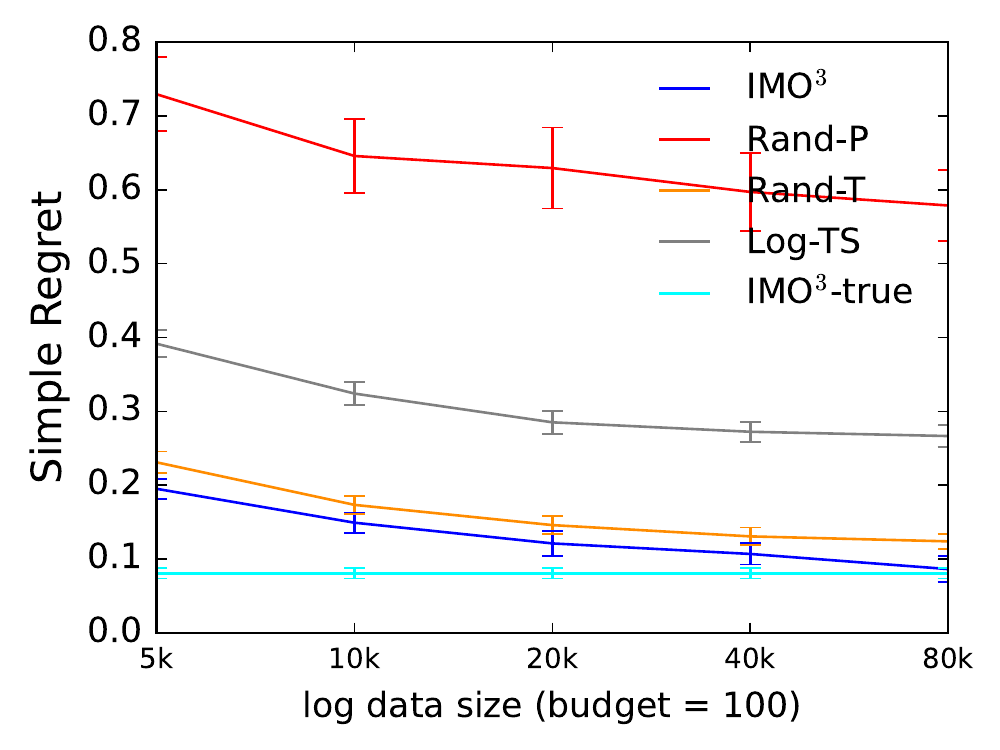}
        \caption{ZDT1.}
        \label{fig:zdt1-varydatasize}
    \end{subfigure}
    \begin{subfigure}[b]{0.245\textwidth}
        \centering
        \includegraphics[width=\linewidth]{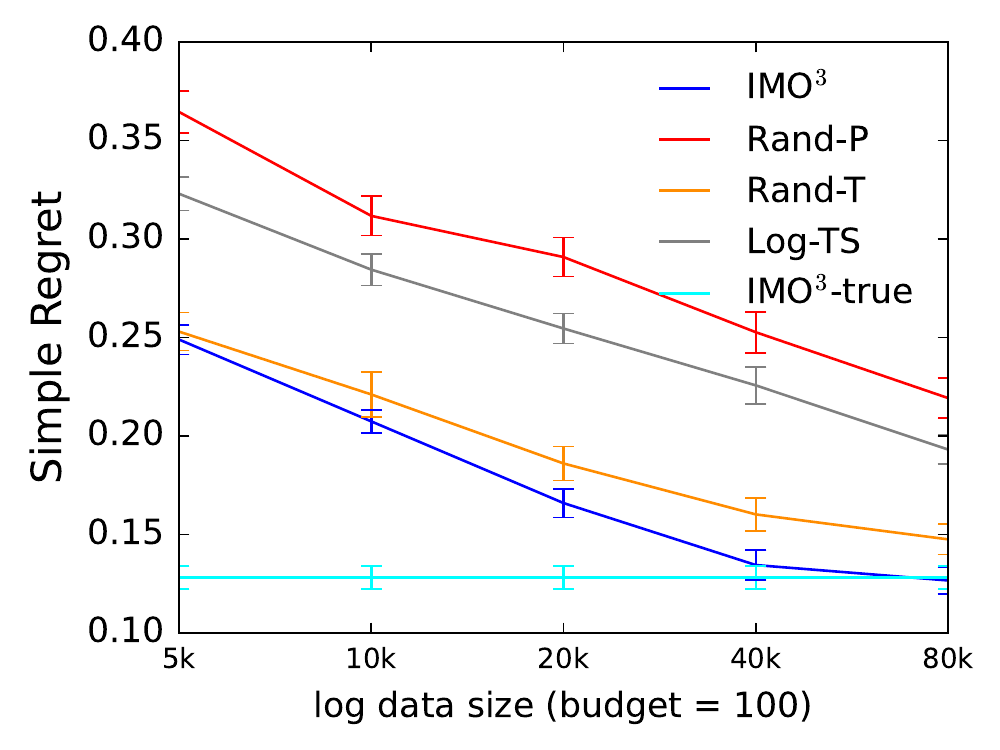}
        \caption{Crashworthiness.}
        \label{fig:crash-varydatasize}
    \end{subfigure}
    \begin{subfigure}[b]{0.245\textwidth}
        \centering
        \includegraphics[width=\linewidth]{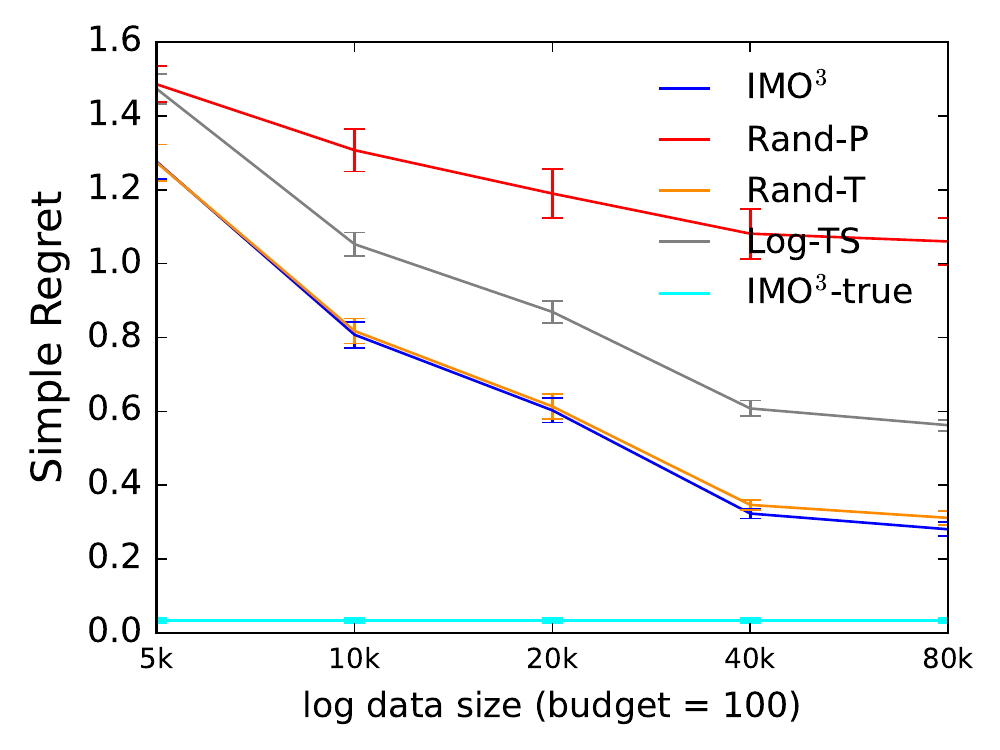}
        \caption{Stock investment.}
        \label{fig:stock-varydatasize}
    \end{subfigure}
    \begin{subfigure}[b]{0.245\textwidth}
        \centering
        \includegraphics[width=\linewidth]{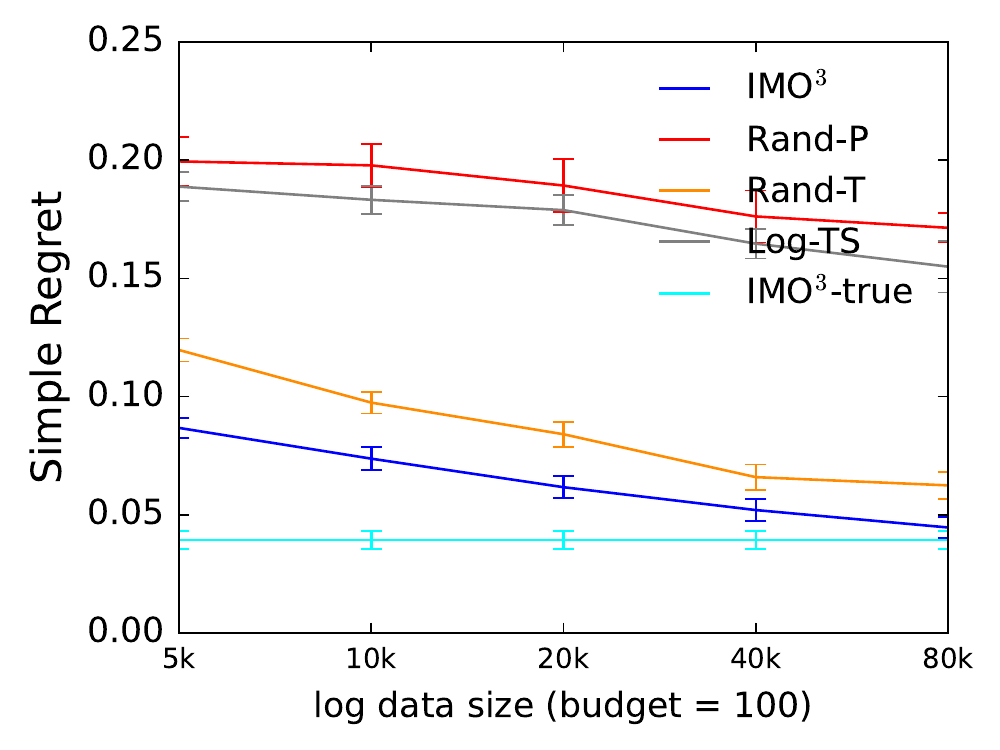}
        \caption{Yahoo! news recommendation.}
        \label{fig:click-varydatasize}
    \end{subfigure}
    \caption{Simple regret of different algorithms by fixing budget $T=100$ and varying logged data size. Each experiment is averaged over 10 logged data, 10 randomly selected $\theta_*$ and 5 runs under each combination of logged data and $\theta_*$.}
\label{fig:varydatasize}
\end{figure*}
\\\noindent
\textbf{Stock Investment.}\hspace*{1mm}
The stock investment problem is a widely studied real-world MOO problem \cite{liang2013portfolio}, where we need to trade off returns 
and volatility of
an investment strategy. We consider investing one dollar in a stock at the end of each day as an action and try to optimize the \emph{relative gain} and \emph{volatility} of this investment at the end of the next day. Specifically, the relative gain is the stock's closing price on the second day minus that on the first day, and we use the \emph{absolute} difference
as a measure of investment volatility.
Our goal is to maximize the relative gain and minimize the volatility between two consecutive days of a one-dollar investment, on average. 
We use 48 popular stocks (see \cref{appendix:exp} for the list) as the action set $\cA$, and the four quarters of a year as the context set $\cX$. We collect the closing stock prices from \href{https://finance.yahoo.com/videos?ncid=dcm_23657983_265755135_460682638_127471542&gclid=Cj0KCQiAoab_BRCxARIsANMx4S5Lsnu5vNLGQXcm_125QN5VdHiPxaXGrXQdM57Y6FF_yHHhVeSd3pIaAlQuEALw_wcB}{Yahoo Finance} for the period Nov.1/2020--Nov.1/2021 for generating logged data. 
\\\noindent
\textbf{Yahoo! News Recommendation.}\hspace*{1mm}
This is a news article recommendation problem derived from the \href{https://webscope.sandbox.yahoo.com/catalog.php?datatype=r}{Yahoo! Today Module click log dataset (R6A)}. We consider two objectives to maximize, the \emph{click through rate (CTR)} and \emph{diversity} of the recommended articles. In the original dataset, each record contains the recommended article, the click event (0 or 1), the pool of candidate articles, and a 6-dimensional feature vector for each article in the pool. The recommended article is selected from the pool uniformly at random. We adopt the original click event in the logged dataset to measure CTR of the recommendation, and use the $\ell_2$ distance between the recommended article's feature and the average feature vector in the pool to represent the diversity of this recommendation. For our experiments, we extract five different article pools as contexts and all logged records associated with them from the original data, resulting in 1,123,158 records in total. Each article pool has 20 candidates as actions. 

\subsection{Baselines}
\label{sec:baselines}
\noindent
\textbf{Random Policy (Rand-P).}\hspace*{1mm}
The \emph{random policy} \cite{arthur2016non-stochastic} is a standard baseline in BAI, which selects a policy (arm) $\pi_t \in \Pi$ uniformly at random from the policy space in each round $t$. The off-policy value estimate $\hat V(\pi_t)$ is presented to the designer for feedback $Y_t$. After $T$ rounds, the value estimates and their feedback are used to form the maximum likelihood estimate of $\theta_*$, $\hat\theta$, which is used to solve \eqref{eq:off-optimization} for the final identified policy.
\\\noindent
\textbf{Random Trade-off (Rand-T).}\hspace*{1mm}
Instead of sampling a random policy, Rand-T samples a trade-off vector $\theta_t$ uniformly at random from a $d$-dimensional unit ball, which is used to identify a policy $\pi_t$ in each round by policy optimization in \eqref{eq:off-optimization}. The rest is the same as the Rand-P baseline. 
\\\noindent
\textbf{Logistic Thompson Sampling (Log-TS).}\hspace*{1mm}
Many cumulative regret minimization algorithms with guarantees exist \cite{marc2017linear,kveton2019randomized}. Therefore, we also consider a \emph{cumulative-to-simple regret reduction} as a baseline. In particular, we adapt Thompson sampling (TS) for generalized linear bandits \cite{marc2017linear,kveton2019randomized} to the BAI problem, and output the ``best'' policy as the average of its selected policies. In each round $t$, we sample a trade-off vector $\theta_t$ from the current posterior over $\theta$ with Log-TS, which is used to identify a policy $\pi_t$ in each round by policy optimization using \eqref{eq:off-optimization}. Then $\hat V(\pi_t)$ and feedback $Y_t$ are used to update the posterior. The final output policy is the average of all policies selected in $T$ rounds, 
$\tilde\pi_*=\sum_{t=1}^T \pi_t / T.$
This reduction of Log-TS leads to a simple regret of $\hat R^{sim}_T = \tilde O(d^{\frac{3}{2}}\sqrt{T\log(1/\delta)})$, where $\tilde O$ stands for the big-O notation up to logarithmic factors in $T$. The proof is in \cref{appendix:exp}.
\\\noindent
\textbf{\imo with different value estimators.}\hspace*{1mm}
We fix the pre-selection budget $L=500$, which requires no designer feedback. To assess the impact of off-policy estimated values on optimization performance, we test variants of \imo with its off-policy estimated values replaced by the true expected values (dubbed \imo-true). We use the IPS estimator by default in this section. Experiments with the DM and DR estimators can be found in \cref{appendix:exp}. 

\subsection{Results and Analysis}
\label{sec:results}
For each of the four problems, we first fix the size of the logged dataset and assess how simple regret ((lower the better)) varies with the interaction budget $T$.
The results are shown in \cref{fig:varybudget}. Each result is averaged over ten logged datasets generated for each problem, ten randomly sampled $\theta_*$, and 5 repeated runs under each combination of the logged data and $\theta_*$ 
(error bars represent standard error). We see that \imo outperforms or performs comparably to our baselines in all four problems. While Rand-T is similar to the pre-selection phase of \imo and performs relatively well, its exploration is less efficient and limited by the budget, and thus is worse than \imo. This illustrates the advantage of using G-optimal design with a sufficient number of pre-selected policies to query the designer for feedback. The gap between \imo using estimated vs.\ true values is due to errors in value estimation---see the second term in our regret bound
(\cref{theorem:regretanalysis}). 
This term is invariant w.r.t.\ $T$, thus the gap remains relatively constant as $T$ varies in our experiments. 

We further study how the \emph{amount} of logged data influences the simple regret of \imo. We fix $T = 100$, and vary the size of the logged dataset used for policy-value estimation. Intuitively, if the dataset is sufficient to provide an accurate value estimate for any policy, \imo should perform similarly to directly using true values. Results
in \cref{fig:varydatasize}
show that when the logged dataset is small, inaccurate value estimates cause algorithms that rely on off-policy estimates to perform poorly compared to using true values. As the size of the dataset increases, the decrease in value-estimation error allows \imo to outperform the baselines by selecting the most effective policies for querying the designer. When the logged dataset is sufficiently large, more accurate value estimates ensure that
\imo converges to the that of using true values.

%% file: related_work.tex
\section{Related Work}
\label{sec:related work}


\citeauthor{drugan13designing} \shortcite{drugan13designing} is the first work to propose, analyze and experiment with a \ucb algorithm with a scalarized objective for MOO and a Pareto \ucb algorithm. \citeauthor{auer16pareto} \shortcite{auer16pareto} formulate the problem of Pareto-frontier identification as a BAI problem.
Thompson sampling in MOO is studied (though not theoretically analyzed) by \citeauthor{yahyaa15thompson} \shortcite{yahyaa15thompson}. 
Two recent works apply Gaussian process (GP) bandits to MOO.
\citeauthor{paria19flexible} \shortcite{paria19flexible} model the posterior of each objective function as a GP and minimize regret w.r.t.\ a known distribution of scalarization vectors. \citeauthor{zhang20random} \shortcite{zhang20random} show that this algorithm generates a set of points that maximize random hypervolume scalarization, a objective often used in practice. All above works are in the online setting, where the learning agent interactively probes the environment to learn about its objective functions. Our setting is offline and the objective functions are estimated from logged data collected by some prior policy.


In terms of the motivation, the closest work to ours is that of \citeauthor{roijers17interactive} \shortcite{roijers17interactive}, 
who treat online MOO as a two-stage problem, where the objective functions are estimated using initial interactions with the environment and the scalarization vector is then estimated via user interaction. Unlike our work, they do not propose a specific algorithm for their setting, but only adapts existing bandit algorithms based on learned utility functions. Besides, they do not formulate the problem as off-policy optimization, and thus the process can be costly. 

%% file: conclusion.tex
\section{Conclusion}
\label{sec:conclusion}

In this work, we study the problem of multi-objective optimization with unknown objective functions. We propose an interactive off-policy optimization algorithm for finding the optimal policy that achieves the desired trade-off among objectives. Specifically, we adapt off-policy estimators to evaluate policy values on all objectives, choose policies that effectively elicit a designer's preference trade-offs, and learn the optimal policy using best arm identification. We prove upper bounds on the simple regret or our method and demonstrate it effectiveness with experiments on four MOO problems.

For future work, we plan to generalize (and analyze) our algorithm to more complex utility functions and other types of query models.
We applied G-optimal design for BAI to provide theoretical guarantees---using other BAI algorithms for MOO is of interest.

%% file: appendix.tex
\section{Off-Policy Optimization}
\label{appendix:linearprog}

\begin{claim}
\label{claim:DM} Using the DM estimator \eqref{eq:DM}, problem \eqref{eq:off-optimization} is maximization of a linear function with linear constraints.
\end{claim}

\noindent The claim is proved as follows. By definition, $\hat{V}^\textsc{dm}(\pi)$ is linear in $\pi$, and so is $\theta\T \hat{V}^\textsc{dm}(\pi)$ for any $\theta$. The set $\Pi$ is an intersection of halfspaces, as discussed below \eqref{eq:optimization}. \qed

\begin{claim}
\label{claim:IPS} Using the IPS estimator \eqref{eq:IPS}, problem \eqref{eq:off-optimization} is maximization of a linear function with linear constraints.
\end{claim}

\noindent Under the assumption that the rewards are non-negative, any clipped policy $\pi$, where $\pi(a \mid x) / \pi_0(a \mid x) > M$ for all $x$ and $a$, can be replaced with an unclipped policy with at least as high value. Let
\begin{equation*}
  \Pi'
  = \set{\pi \in \Pi: \pi(a \mid x) \leq M \pi_0(a \mid x), \ \forall x \in \cX, a \in \cA}
\end{equation*} 
be the set of all unclipped policies. This set has two key properties. First, it is an intersection of halfspaces, since $\Pi$ is and the additional constraints are linear in $\pi$. Second, for any $\pi \in \Pi'$, the minimum in \eqref{eq:IPS} can be omitted. In turn, $\hat{V}^\textsc{ips}(\pi)$ becomes linear in $\pi$ and so does $\theta\T \hat{V}^\textsc{ips}(\pi)$ for any $\theta$. \qed

\begin{claim}
\label{claim:DR} Using the DR estimator \eqref{eq:DR}, the problem \eqref{eq:off-optimization} is maximization of a linear function with linear constraints.
\end{claim}

\noindent  The claim is proved as follows. By definition, $\hat{V}^\textsc{dr}(\pi)$ is linear in $\pi$, and so is $\theta\T \hat{V}^\textsc{dr}(\pi)$ for any $\theta$. The set $\Pi$ is an intersection of halfspaces, as discussed below \eqref{eq:optimization}. \qed

\section{Regret Analysis}
\label{appendix:proof}

This section is organized as follows. \cref{theorem:generalbound} gives a general simple regret bound of a BAI algorithm for MOO, which decomposes into the simple regret of the algorithm based on estimated policy values, and a second term that accounts for errors in off-policy estimated values. In \cref{theorem:offsimpleregret}, we bound the simple regret of \imo based on off-policy estimated values. In \cref{theorem:ipserror}, we bound the error induced by the IPS estimator.

\begin{theorem}
\label{theorem:generalbound} For any policy $\pi \in \Pi$, let $||\hat V(\pi) - V(\pi)||_2 \leq \varepsilon$ hold with probability at least $1 - \delta$. Then the simple regret of \imo is
\begin{equation*}
  R_T^{sim}
  \leq \hat R_T^{sim} + 2\varepsilon||\theta_*||_2
\end{equation*}
with probability at least $1-2\delta$, where $\hat R_T^{sim} = u_{\theta_*}(\hat V(\hat\pi_*)) - u_{\theta_*}(\hat V(\tilde\pi_*))$ and $\hat\pi_*=\argmax_{\pi\in\Pi} u_{\theta_*}\big(\hat V(\pi)\big)$ are the simple regret and optimal policy with respect to the estimated policy values, respectively.
\end{theorem}
\begin{proof}
\label{proof:generalbound}
First recall that $\pi_\ast = \argmax_{\pi\in\Pi} \theta_\ast\T V(\pi)$ is the optimal policy with respect to \emph{true values}, $\hat\pi_\ast = \argmax_{\pi\in\Pi} \theta_\ast\T\hat V(\pi)$ is the optimal policy with respect to \emph{estimated values}, and $\tilde\pi_\ast$ is the output of the \imo algorithm.
The simple regret is 
\begin{align*}
  R_T^{sim}
  & = \theta_\ast^\top V(\pi_*) - \theta_\ast^\top V(\tilde\pi_*) \\
  & = \theta_\ast^\top V(\pi_*) - \theta_\ast^\top \hat V(\hat\pi_*) + \theta_\ast^\top \hat V(\hat\pi_*) - \theta_\ast^\top V(\tilde\pi_*) \\
  & \leq \theta_\ast^\top V(\pi_*) - \theta_\ast^\top \hat V(\pi_*) + \theta_\ast^\top \hat V(\pi_*) - \theta_\ast^\top \hat V(\hat\pi_*) + \theta_\ast^\top \hat V(\hat\pi_*) - \theta_\ast^\top \hat V(\tilde\pi_*) + \theta_\ast^\top \hat V(\tilde\pi_*) - \theta_\ast^\top V(\tilde\pi_*)\,.
\end{align*} 
Note that $\hat R_T^{sim} = \theta_\ast^\top \hat V(\hat\pi_*) - \theta_\ast^\top \hat V(\tilde\pi_*)$ and $\theta_\ast^\top \hat V(\pi_*) - \theta_\ast^\top \hat V(\hat\pi_*)\leq 0$. Therefore,
\begin{align*}
  R_T^{sim}
  & \leq \hat R_T^{sim} + \theta_\ast^\top V(\pi_*) - \theta_\ast^\top \hat V(\pi_*) + \theta_\ast^\top \hat V(\tilde\pi_*) - \theta_\ast^\top V(\tilde\pi_*) \\
  & \leq \hat R_T^{sim} + |\theta_\ast^\top V(\pi_*) - \theta_\ast^\top \hat V(\pi_*)| + |\theta_\ast^\top \hat V(\tilde\pi_*) - \theta_\ast^\top V(\tilde\pi_*)| \\
  & \leq \hat R_T^{sim} + ||\theta_\ast||_2 ||V(\pi_*) - \hat V(\pi_*)||_2 + ||\theta_\ast||_2 ||\hat V(\tilde\pi_*) - V(\tilde\pi_*)||_2\,,
\end{align*}
where the last step is by the Cauchy-Schwarz inequality. Finally, $||\hat V(\pi) - V(\pi)||_2 \leq \varepsilon$ holds for any policy $\pi$ with probability at least $1-\delta$, and thus
\begin{align*}
  R_T^{sim}
  \leq \hat R_T^{sim} + 2\varepsilon||\theta_\ast||_2
\end{align*}
holds with probability at least $1 - 2\delta$ by the union bound. This concludes the proof.
\end{proof}

The general bound in \cref{theorem:generalbound} decomposes into two parts. The first term $\hat R_T^{sim}$ is the regret of BAI based on \emph{estimated policy values} and reflects the amount of the designer's feedback. We bound it for \imo in \cref{theorem:offsimpleregret}. The second term accounts for errors in off-policy estimates. We bound the off-policy error $\varepsilon$ in \cref{theorem:ipserror}. Both theorems are stated and proved below.

\begin{theorem}
\label{theorem:offsimpleregret} Let $c_{\min}, \delta > 0$ be chosen such that
\begin{align*}
  \min_{v \in \cV} \min \{h'(v\T \theta_*), h'(v\T \hat{\theta})\}
  \geq c_{\min}
\end{align*}
holds with probability at least $1 - \delta$. Then
\begin{align*}
  \hat{R}_T^{sim}
  \leq L \exp\left[- \frac{\Delta_{\min}^2 c_{\min}^2 T}{2 g(\alpha_*)}\right]
\end{align*}
holds with probability at least $1 - \delta$.
\end{theorem}
\begin{proof}
Let $\hat{\theta}$ be the MLE of model parameters returned by \imo, estimated from a dataset of size $T$ collected according to the optimal design $\alpha_*$. Let $\hat{\mu}_i = v_i\T \hat{\theta}$ be the corresponding utility estimate. To simplify exposition, we do not analyze the effect of rounding in the optimal design and assume that all $\alpha_*(i)$ are multiples of $1 / T$. In this case, each $v_i$ appears in the collected dataset exactly $\alpha_*(i) T$ times. This is a standard assumption in the analyses with optimal designs.

Now we are ready to bound the simple regret of our solution. Let $I$ be the index of the policy chosen by our algorithm. Then
\begin{align*}
  \hat{R}_T^{sim}
  & = \E{\mu_1 - \mu_I}
  \leq \prob{I > 1}
  \leq \sum_{i = 2}^L \prob{\hat{\mu}_i \geq \hat{\mu}_1}
  \leq \sum_{i = 2}^L (\prob{\hat{\mu}_i \geq \mu_i + \Delta_i / 2} +
  \prob{\hat{\mu}_1 \leq \mu_1 - \Delta_i / 2}) \\
  & = \sum_{i = 2}^L (\prob{\hat{\mu}_i - \mu_i \geq \Delta_i / 2} +
  \prob{\mu_1 - \hat{\mu}_1 \geq \Delta_i / 2})\,.
\end{align*}
The first inequality follows from $\mu_i \in [0, 1]$ and the second is a result of applying a union bound over all policies. In the third inequality, we use that $a \geq b$ implies that either $a \geq c$ or $b \leq c$ holds for any $c$, which we choose as $c = (\mu_i + \mu_1) / 2$.

By Lemma 1 in \cite{kveton2019randomized}, the MLE $\hat{\theta}$ in a GLM satisfies
\begin{align*}
  \hat{\theta} - \theta_*
  = \Lambda^{-1} \sum_{t = 1}^T X_t \varepsilon_t\,,
\end{align*}
where $X_t = \hat{V}(\pi_t)$ are estimated values in round $t$ and $\varepsilon_t$ is observation noise. Note that the noise is $\sigma^2$-sub-Gaussian for $\sigma = 1 / 2$. Moreover, $\Lambda = \sum_{t = 1}^T h'(X_t\T \tilde{\theta}) X_t X_t\T$ is the sample covariance matrix weighted by the derivative of $h$ at $\tilde{\theta}$, a convex combination of $\hat{\theta}$ and $\theta_*$. After this decomposition, we can apply Hoeffding's inequality to a sum of weighted sub-Gaussian random variables and get
\begin{align*}
  \prob{\hat{\mu}_i - \mu_i \geq \Delta_i / 2}
  & = \prob{v_i\T (\hat{\theta} - \theta_*) \geq \Delta_i / 2}
  \leq \exp\left[- \frac{\Delta_i^2}{8 \sigma^2 v_i\T \Lambda^{-1} G \Lambda^{-1} v_i}\right]
  = \exp\left[- \frac{\Delta_i^2}{2 v_i\T \Lambda^{-1} G \Lambda^{-1} v_i}\right]
\end{align*}
for any $i \in [L]$, where $G = \sum_{t = 1}^T X_t X_t\T$.

Now we make three observations. First, since $\tilde{\theta}$ is a convex combination of $\hat{\theta}$ and $\theta_*$, and the derivative of the logistic function is monotone, we have $h'(X_t\T \tilde{\theta}) \geq c_{\min}$. Second, since $h'(X_t\T \tilde{\theta}) \geq c_{\min}$, we have $c_{\min} G \preceq \Lambda$ and thus $c_{\min}^{-1} G^{-1} \succeq \Lambda^{-1}$. Third, since our optimal design is applied exactly, $G = T G_{\alpha_*}$. It follows that
\begin{align*}
  v_i\T \Lambda^{-1} G \Lambda^{-1} v_i
  \leq c_{\min}^{-2} v_i\T G^{-1} v_i
  = c_{\min}^{-2} T^{-1} v_i\T G_{\alpha_*}^{-1} v_i\,,
\end{align*}
and in turn
\begin{align*}
  \prob{\hat{\mu}_i - \mu_i \geq \Delta_i / 2}
  \leq \exp\left[- \frac{\Delta_i^2 c_{\min}^2 T}{2 v_i\T G_{\alpha_*}^{-1} v_i}\right]
  \leq \exp\left[- \frac{\Delta_i^2 c_{\min}^2 T}{2 g(\alpha_*)}\right]\,.
\end{align*}
Finally, we chain all inequalities and get
\begin{align*}
  \hat{R}_T^{sim}
  \leq 2 \sum_{i = 1}^L \exp\left[- \frac{\Delta_i^2 c_{\min}^2 T}{2 g(\alpha_*)}\right]
  \leq L \max_{i \in [L]} \exp\left[- \frac{\Delta_i^2 c_{\min}^2 T}{2 g(\alpha_*)}\right]
  = L \exp\left[- \frac{\Delta_{\min}^2 c_{\min}^2 T}{2 g(\alpha_*)}\right]\,.
\end{align*}
This concludes the proof.
\end{proof}

\begin{theorem}
\label{theorem:ipserror} For any fixed policy $\pi$ and $\delta\in (0,1)$, the error in the value estimate by the IPS estimator is $||\hat V^\textsc{IPS}(\pi) - V(\pi)||_2 \leq \sqrt{\frac{d M^2 \log(2 d /\delta)}{2 N}}$ with probability at least $1-\delta$, where $d$ is the number of objectives, $M$ is the tunable parameter in the IPS estimator, and $N$ is the size of logged data.
\end{theorem}
\begin{proof}
For any objective $i \in [d]$, we can apply Hoeffding's inequality to the random variables $X_j = \min \Big\{M, \frac{\pi(a_j \mid x_j)}{\pi_0(a_j \mid x_j)}\Big\} r_j$ and get that $|\hat V_i^\textsc{ips}(\pi) - V_i(\pi)| \leq M \sqrt{\frac{\log(2 d /\delta)}{2 N}}$ holds with probability at least $1-\delta / d$ \cite{strehl2010learning}. By the union bound,
\begin{align*}
  ||\hat V^\textsc{IPS}(\pi) - V(\pi)||_2
  = \sqrt{\sum_{i = 1}^d (\hat V_i^\textsc{ips}(\pi) - V_i(\pi))^2}
  \leq \sqrt{\frac{d M^2 \log(2 d /\delta)}{2 N}}
\end{align*}
holds with probability at least $1 - \delta$. This concludes the proof.
\end{proof}

Finally, we substitute the bounds from \cref{theorem:offsimpleregret,theorem:ipserror} to \cref{theorem:generalbound}, which leads to \cref{theorem:regretanalysis}.

\section{Experiments}
\label{appendix:exp}

\subsection{Multi-Objective Optimization Problems}
\noindent
\textbf{ZDT1.}\hspace*{1mm}
The ZDT test suite \cite{zitzler2000comparison} is the most widely employed benchmark for MOO.
We use ZDT1, the first problem in the test suite, a box-constrained $n$-dimensional two-objective problem, with objectives $F_1$ and $F_2$ defined as
\begin{align}
    & F_1(x) = 5x_1, \quad F_2(x) = g(x) \bigg[1-\sqrt{\frac{x_1}{g(x)}}\bigg]\,, \\\nonumber
    & \textrm{and }\; g(x) = 1 + \frac{9(\sum_{i=2}^n x_i)}{n-1}\,,
\end{align}
where $x=(x_i)_{i=1}^n$ and $x_i\in [0,1], \forall i\in[n]$. We use $n=5$ in our experiments, treating $(x_4, x_5)$ as context, and perform optimization on $(x_i)_{i=1}^3$. We sample five combinations of $(x_4, x_5)$ uniformly to create $\cX$ and ten combinations of $(x_i)_{i=1}^3$ to create the action set $\cA$. For each context $x\in\cX$, the logging policy is a distribution sampled from a Dirichlet $\pi_0(\cdot \mid x) \sim \mathrm{Dir}(\alpha)$ ($\alpha$ is a $10$-D vector, $\alpha_i=10, \forall i\in[10]$). To generate a logged record in $\mathcal D$, we randomly select a context $x$, use $\pi_0(\cdot \mid x)$ to select an action, and generate its 2-D reward using $F_1$ and $F_2$ with added zero-mean Gaussian noise $N(0, 0.5^2)$. 
\\\noindent
\textbf{Crashworthiness.}\hspace*{1mm}
This MOO problem is extracted from a real-world crashworthiness domain \cite{carvalho2018solvingRM}, where three objectives factor into the optimization of the crash-safety level of a vehicle. We refer to Sec.~2.1 of \cite{carvalho2018solvingRM} for detailed objective functions and constraints. We omit the constants in their objective functions to ensure the three objectives lie in a similar range (if one objective dominates the others, the problem may reduce to a single objective problem). Five bounded decision variables $(x_i)_{i=1}^5$  represent the thickness of reinforced members around the car front. We  use the last two variables as contexts and the first three as actions. The rest settings of the simulation are the same as for ZDT1. 
\\\noindent
\textbf{Stock Investment.}\hspace*{1mm}
The stock investment problem is a widely studied real-world MOO problem \cite{liang2013portfolio}, where we need to trade off returns and volatility of an investment strategy. We consider investing one dollar in a stock at the end of each day as an action and try to optimize the \emph{relative gain} and \emph{volatility} of this investment at the end of the next day. Specifically, the relative gain is the stock's closing price on the second day minus that on the first day, and we use the \emph{absolute} difference
as a measure of investment volatility.
Our goal is to maximize the relative gain and minimize the volatility between two consecutive days of a one-dollar investment, on average. 

We use 48 popular stocks, including \textit{CSCO, UAL, BA, BBY, BAC, LYFT, PEP, COST, LOW, SBUX, AMZN, INTC, GM, ATT, KO, MSFT, UBER, AMD, PINS, NVDA, BBBY, FDX, AXP, FB, IBM, WFC, GS, DELL, NFLX, JPM, COF, MRNA, TSLA, BYND, AAL, JD, GOOG, PFE, FORD, MS, ZM, DAL, BABA, MA, TGT, AAPL, WMT, CRM} as the action set $\cA$. We use the four quarters of a year as the context set $\cX$. To create logged data, we first collect the closing stock prices from \href{https://finance.yahoo.com/videos?ncid=dcm_23657983_265755135_460682638_127471542&gclid=Cj0KCQiAoab_BRCxARIsANMx4S5Lsnu5vNLGQXcm_125QN5VdHiPxaXGrXQdM57Y6FF_yHHhVeSd3pIaAlQuEALw_wcB}{Yahoo Finance} for the period Nov.1/2020--Nov.1/2021. For each context $x\in\cX$, the logging policy is a distribution sampled from a Dirichlet $\pi_0(\cdot \mid x)\sim \mathrm{Dir}(\alpha)$ ($\alpha$ is a $48$-D vector with $\alpha_i=10, \forall i\in[48]$). To generate each logged record, we uniformly sample a context/quarter $x$, sample an action/stock for investment from $\pi_0(\cdot \mid x)$, and sample the two-D reward vector by randomly choosing two consecutive days in the quarter and the selected stock's closing prices on these days to compute relative gain and volatility.  We estimate the value $\hat V(\pi)$ of policy $\pi$ using this logged data and compute its true expected value $V(\pi)$ using the original data. 
\\\noindent
\textbf{Yahoo! News Recommendation.}\hspace*{1mm}
This is a news article recommendation problem derived from the \href{https://webscope.sandbox.yahoo.com/catalog.php?datatype=r}{Yahoo! Today Module click log dataset (R6A)}. We consider two objectives to maximize, the \emph{click through rate (CTR)} and \emph{diversity} of the recommended articles. In the original dataset, each record contains the recommended article, the click event (0 or 1), the pool of candidate articles, and a 6-dimensional feature vector for each article in the pool. The logged recommendation is selected from the pool uniformly. We adopt the original click event in the logged dataset to measure CTR of the recommendation, and use the $\ell_2$ distance between the recommended article's feature and the average feature vector in the pool to represent the diversity of this recommendation. 

For our experiments, we extract five different article pools as contexts and all logged records associated with them from the original data, resulting in 1,123,158 records in total. Each article pool has 20 candidates as actions. To generate a recommendation record in the logged data of certain size, we first randomly sample an article pool as the context, and then sample a record from the original data associated with this article pool. Note that in this way we inherit the uniform logging policy of the original data. 
To aid visualization, we multiply CTR and diversity by 10, so estimated values are not too small. We add zero-mean Gaussian noise ($N(0,  0.5^2)$) to the diversity of each logged recommendation to introduce observational noise. The true policy value is estimated using the full, original dataset without added noise.

\subsection{Regret Analysis of Log-TS for BAI}
\begin{restatable}[]{lemma}{regretaverage}
\label{lemma:regret aver} For any $\delta \in (0, 1)$ and the average policy $\tilde\pi_*=\sum_{t=1}^T \pi_t / T$ from Log-TS, the simple regret is $\hat R_T^{sim} = \tilde{O}(d^\frac{3}{2} \sqrt{\log(1 / \delta) / T})$ with probability at least $1 - \delta$.
\end{restatable}
\begin{proof}
\label{proof:regret aver}
Let $\mu(x) = 1 / (1 + \exp[- x])$ be the sigmoid function. Let $v_* = \hat{V}(\hat{\pi}_*)$ be the optimal arm under estimated policy values and $V_t = \hat{V}(\pi_t)$ be the arm pulled by \imo in round $t \in [T]$. Then, from \cite{marc2017linear}, the expected $n$-round regret of TS in a logistic bandit, which is \imo in this case, is
\begin{align*}
  T \mu(\theta_*\T v_*) - \sum_{t = 1}^T \mu(\theta_*\T V_t)
  = \tilde{O}(d^\frac{3}{2} \sqrt{T \log(1 / \delta)})
\end{align*}
with probability at least $1 - \delta$, for any $\delta\in (0,1)$. Now note that $x - y \leq \alpha^{-1} (\mu(x) - \mu(y))$ holds for any $x > y$, where $\alpha = \min \set{\dot{\mu}(x), \dot{\mu}(y)}$ is the minimum derivative of $\mu$ at $x$ and $y$. Since $\normw{\theta_*}{2}$, $\normw{v_*}{2}$, and $\normw{v_t}{2}$ are bounded, $\alpha$ is bounded away from zero and thus $\alpha^{-1}$ is bounded away from infinity. Thus $\alpha$ can be treated as a constant and
\begin{align*}
  T \theta_*\T v_* - \sum_{t = 1}^T \theta_*\T V_t
  = \tilde{O}(d^\frac{3}{2} \sqrt{T \log(1 / \delta)})\,.
\end{align*}
Finally we note that
\begin{align*}
  \hat R_T^{sim}
  = \frac{1}{T} T \theta_*\T v_* - \frac{1}{T} \sum_{t = 1}^T \theta_*\T V_t
  = \tilde{O}(d^\frac{3}{2} \sqrt{\log(1 / \delta) / T})\,.
\end{align*}
\end{proof}

\subsection{\imo with Different Off-Policy Estimators}
Besides the IPS estimator, we also apply DM and DR estimators in \imo, to show how \imo performs with different off-policy estimators. We only evaluate DM and DR estimators on the first three problems. As the logging policy used to collect the  original Yahoo! Today Module click data is already a uniform policy, DM and DR estimators are exactly the same as the IPS estimator. For each of the first three problems in \cref{sec:exp}, we first build a reward model learned from the logged data, and then apply the model in DM and DR estimators. Specifically, for the DM estimator, we simply use the empirical mean of an action as its estimated value.

\begin{figure*}[ht]
    \centering
    \begin{subfigure}[b]{0.32\textwidth}
        \centering
        \includegraphics[width=\linewidth]{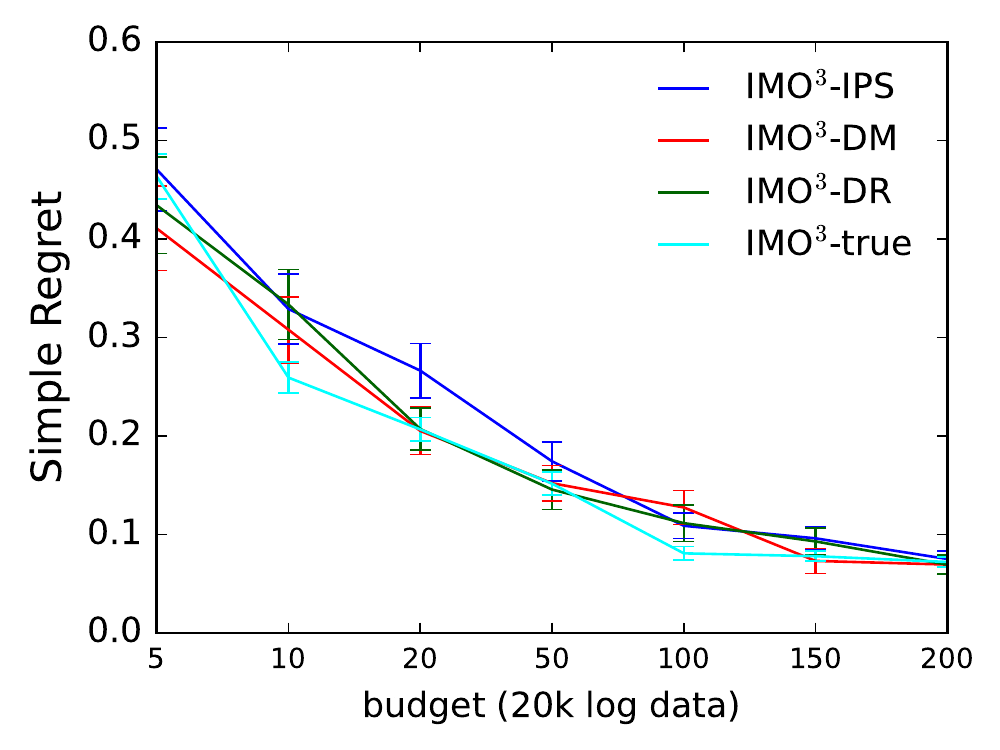}
        \caption{ZDT1.}
        \label{fig:estimator_zdt1-varybudget}
    \end{subfigure}
    \begin{subfigure}[b]{0.32\textwidth}
        \centering
        \includegraphics[width=\linewidth]{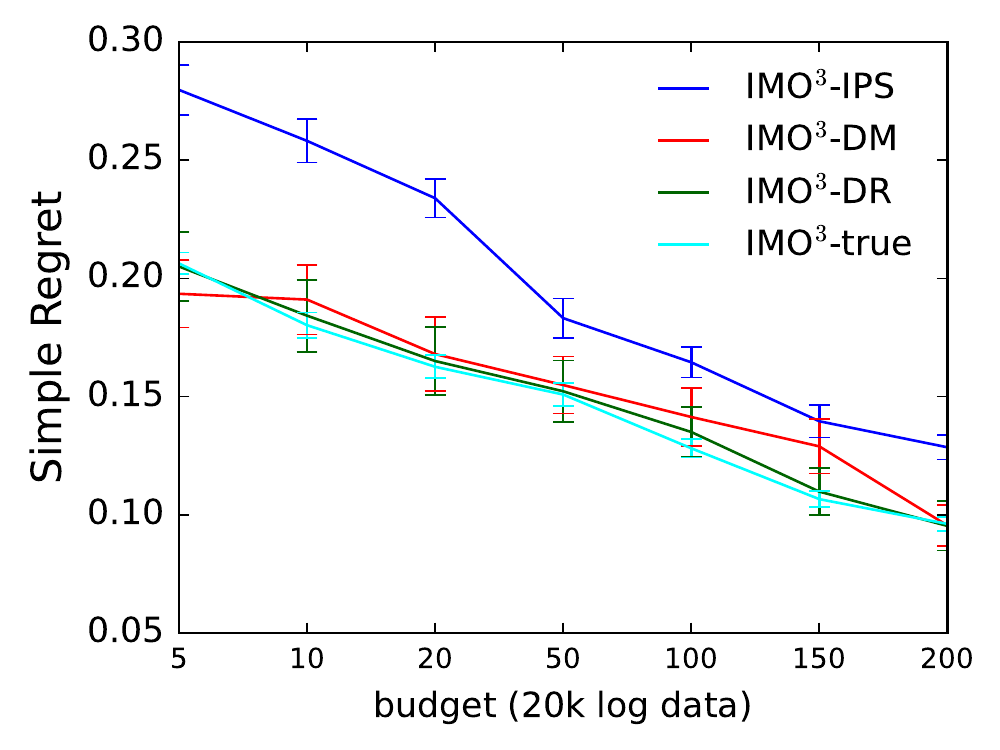}
        \caption{Crashworthiness.}
        \label{fig:estimator_crash-varybudget}
    \end{subfigure}
    \begin{subfigure}[b]{0.32\textwidth}
        \centering
        \includegraphics[width=\linewidth]{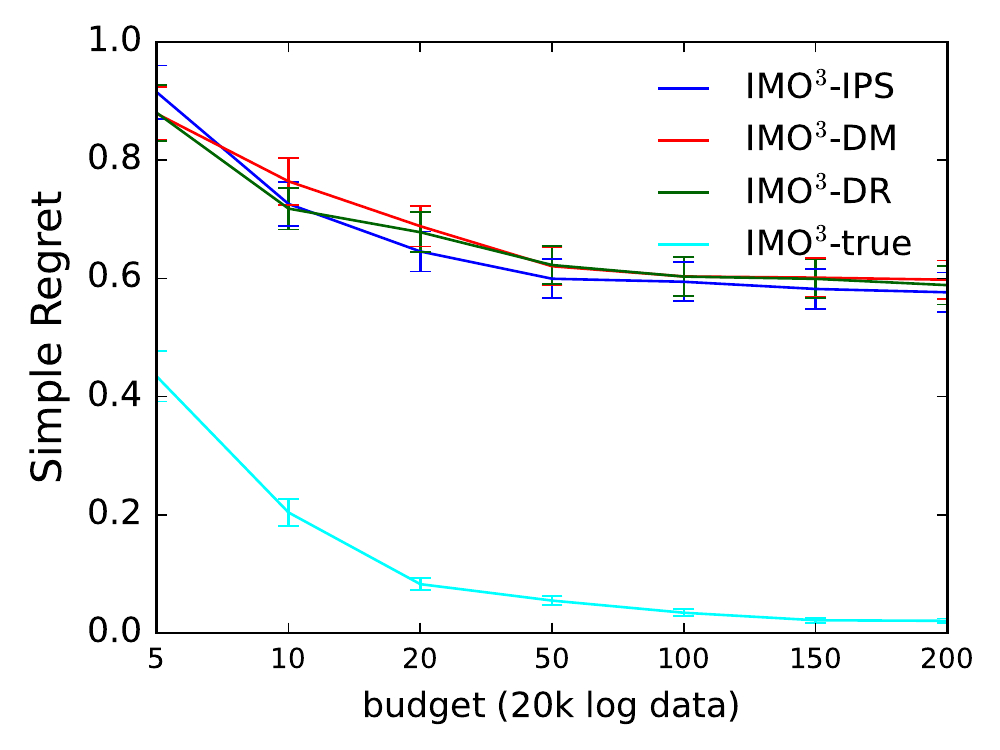}
        \caption{Stock investment. }
        \label{fig:estimator_stock-varybudget}
    \end{subfigure}
    \caption{Simple regret of different algorithms by fixing logged data size $N=20,000$ and varying budget. Each experiment is averaged over 10 logged data, 10 randomly selected $\theta_*$ and 5 runs under each combination of logged data and $\theta_*$.}
\label{fig:estimator_varybudget}
\end{figure*}
\begin{figure*}[ht]
    \centering
    \begin{subfigure}[b]{0.32\textwidth}
        \centering
        \includegraphics[width=\linewidth]{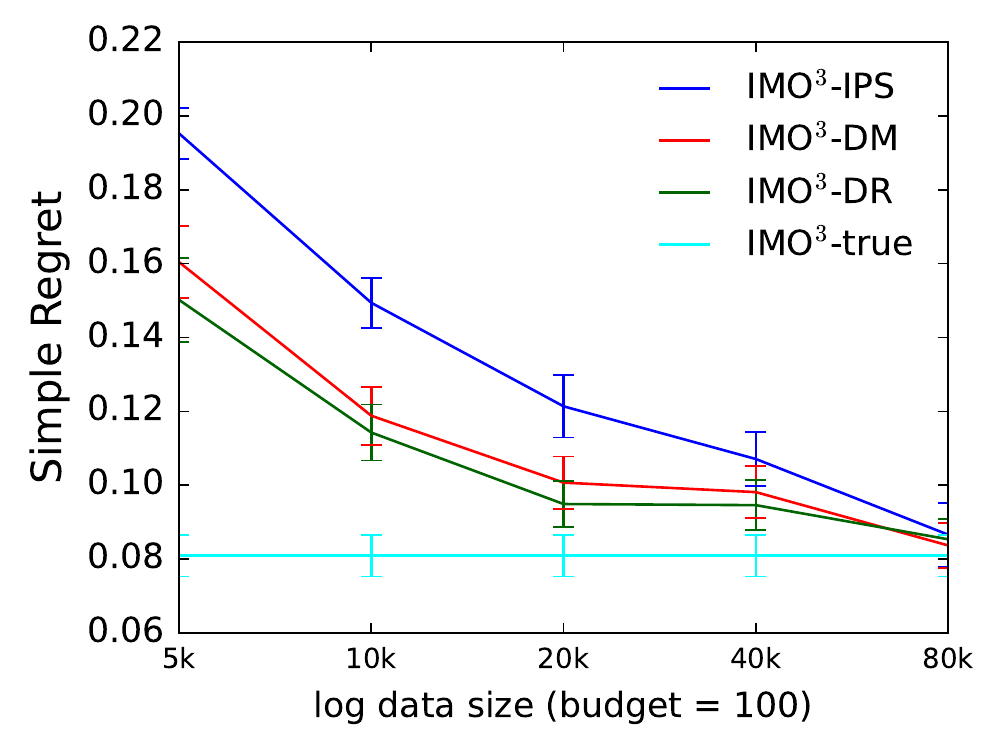}
        \caption{ZDT1.}
        \label{fig:estimator_zdt1-varydatasize}
    \end{subfigure}
    \begin{subfigure}[b]{0.32\textwidth}
        \centering
        \includegraphics[width=\linewidth]{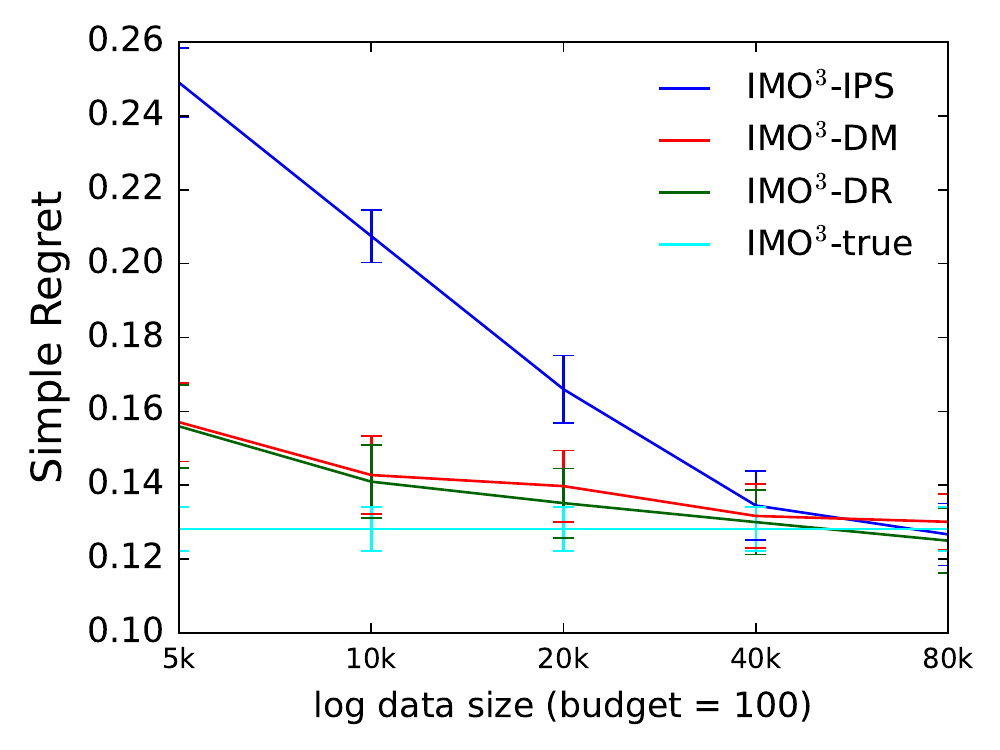}
        \caption{Crashworthiness.}
        \label{fig:estimator_crash-varydatasize}
    \end{subfigure}
    \begin{subfigure}[b]{0.32\textwidth}
        \centering
        \includegraphics[width=\linewidth]{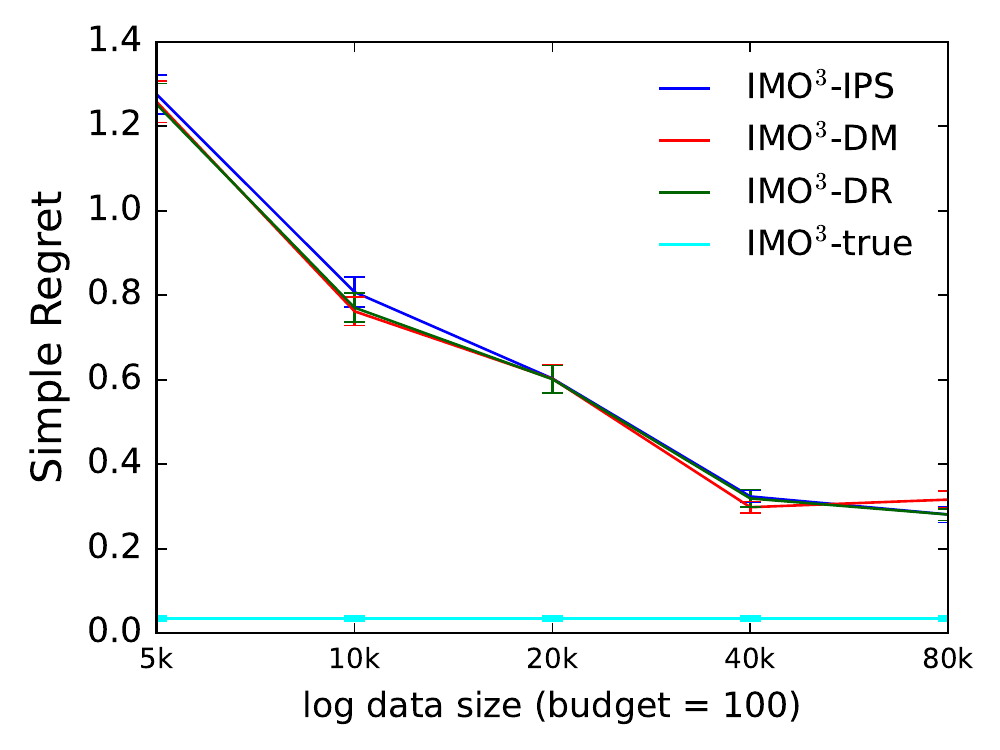}
        \caption{Stock investment.}
        \label{fig:estimator_stock-varydatasize}
    \end{subfigure}
    \caption{Simple regret of different algorithms by fixing budget $T=100$ and varying logged data size. Each experiment is averaged over 10 logged data, 10 randomly selected $\theta_*$ and 5 runs under each combination of logged data and $\theta_*$.}
\label{fig:estimator_varydatasize}
\end{figure*}

We follow the experimental setting in \cref{sec:results}, and show the performance of the three estimators in \cref{fig:estimator_varybudget,fig:estimator_varydatasize}. We observe that \imo performs consistently well with all three estimators, which demonstrates the robustness of our method. Besides, the DM and DR estimators can achieve comparable or even better performance than the IPS estimator across different problems, which may be due to the higher variance of IPS. In particular, when the logging policy used to generate logged data has small probabilities on certain actions, it can lead to high variance in the value estimates. The DR estimator is usually slightly better than the other two, which demonstrates the improved performance with a better off-policy estimator.

%% file: paper.bbl
\begin{thebibliography}{}

\bibitem[\protect\citeauthoryear{Abeille and Lazaric}{2017}]{marc2017linear}
Marc Abeille and Alessandro Lazaric.
\newblock Linear thompson sampling revisited.
\newblock In {\em AISTATS}, 2017.

\bibitem[\protect\citeauthoryear{Auer \bgroup \em et al.\egroup
  }{2016}]{auer16pareto}
Peter Auer, Chao{-}Kai Chiang, Ronald Ortner, and Madalina~M. Drugan.
\newblock Pareto front identification from stochastic bandit feedback.
\newblock In {\em AISTATS}, 2016.

\bibitem[\protect\citeauthoryear{Beygelzimer and
  Langford}{2009}]{beygelzimer2016offset}
Alina Beygelzimer and John Langford.
\newblock The offset tree for learning with partial labels.
\newblock KDD, 2009.

\bibitem[\protect\citeauthoryear{Boutilier}{2002}]{preference:aaai02}
Craig Boutilier.
\newblock A {POMDP} formulation of preference elicitation problems.
\newblock 2002.

\bibitem[\protect\citeauthoryear{Boutilier}{2013}]{boutilier:regretSurvey2013}
Craig Boutilier.
\newblock Computational decision support: Regret-based models for optimization
  and preference elicitation.
\newblock In {\em Comparative Decision Making: Analysis and Support Across
  Disciplines and Applications}. 2013.

\bibitem[\protect\citeauthoryear{Branke \bgroup \em et al.\egroup
  }{2008}]{branke2008multiobjective}
J.~Branke, K.~Deb, Kaisa Miettinen, and R.~Slowinski.
\newblock {\em Multiobjective Optimization: Interactive and Evolutionary
  Approaches}.
\newblock Springer-Verlag, 2008.

\bibitem[\protect\citeauthoryear{Camerer}{2004}]{camerer:2003}
Colin~F. Camerer.
\newblock {\em Advances in Behavioral Economics}.
\newblock 2004.

\bibitem[\protect\citeauthoryear{de Carvalho \bgroup \em et al.\egroup
  }{2018}]{carvalho2018solvingRM}
Vinicius~Renan de~Carvalho, Jaime Sim{\~a}o, and Sichman.
\newblock Solving real-world multi-objective engineering optimization problems
  with an election-based hyper-heuristic.
\newblock 2018.

\bibitem[\protect\citeauthoryear{Deaton and
  Cartwright}{2018}]{deaton2018randomized}
Angus Deaton and Nancy Cartwright.
\newblock Understanding and misunderstanding randomized controlled trials.
\newblock {\em Social Science \& Medicine}, 2018.

\bibitem[\protect\citeauthoryear{Drugan and
  Now{\'{e}}}{2013}]{drugan13designing}
Madalina~M. Drugan and Ann Now{\'{e}}.
\newblock Designing multi-objective multi-armed bandits algorithms: {A} study.
\newblock In {\em Proceedings of the 2013 International Joint Conference on
  Neural Networks}, 2013.

\bibitem[\protect\citeauthoryear{Dudik \bgroup \em et al.\egroup
  }{2011}]{dudik2011doubly}
Miroslav Dudik, John Langford, and Lihong Li.
\newblock Doubly robust policy evaluation and learning.
\newblock {\em ICML}, 2011.

\bibitem[\protect\citeauthoryear{Dudík \bgroup \em et al.\egroup
  }{2015}]{miroslav2015contextual}
Miroslav Dudík, Katja Hofmann, Robert~E. Schapire, Aleksandrs Slivkins, and
  Masrour Zoghi.
\newblock Contextual dueling bandits.
\newblock In {\em COLT}, 2015.

\bibitem[\protect\citeauthoryear{Jamieson and
  Talwalkar}{2016}]{arthur2016non-stochastic}
Kevin Jamieson and Ameet Talwalkar.
\newblock Non-stochastic best arm identification and hyperparameter
  optimization.
\newblock In {\em AISTATS}. PMLR, 2016.

\bibitem[\protect\citeauthoryear{Karnin \bgroup \em et al.\egroup
  }{2013}]{karnin2013almost}
Zohar Karnin, Tomer Koren, and Oren Somekh.
\newblock Almost optimal exploration in multi-armed bandits.
\newblock In {\em ICML}, 2013.

\bibitem[\protect\citeauthoryear{Keeney and Raiffa}{1976}]{keeney-raiffa}
Ralph~L. Keeney and Howard Raiffa.
\newblock {\em Decisions with Multiple Objectives: Preferences and Value
  Trade-offs}.
\newblock Wiley, 1976.

\bibitem[\protect\citeauthoryear{Kohavi and
  Longbotham}{2011}]{kohavi2011unexpected}
Ron Kohavi and Roger Longbotham.
\newblock Unexpected results in online controlled experiments.
\newblock {\em SIGKDD Explor. Newsl.}, 2011.

\bibitem[\protect\citeauthoryear{Kohavi \bgroup \em et al.\egroup
  }{2009}]{kohavi2009controlled}
Ron Kohavi, Roger Longbotham, Dan Sommerfield, and Randal~M. Henne.
\newblock Controlled experiments on the web: Survey and practical guide.
\newblock {\em KDD}, 2009.

\bibitem[\protect\citeauthoryear{Kveton \bgroup \em et al.\egroup
  }{2020}]{kveton2019randomized}
Branislav Kveton, Manzil Zaheer, Csaba Szepesvari, Lihong Li, Mohammad
  Ghavamzadeh, and Craig Boutilier.
\newblock Randomized exploration in generalized linear bandits.
\newblock In {\em AISTATS}, 2020.

\bibitem[\protect\citeauthoryear{Lambert and Pregibon}{2007}]{diane2007more}
Diane Lambert and Daryl Pregibon.
\newblock More bang for their bucks: assessing new features for online
  advertisers.
\newblock {\em {SIGKDD} Explor.}, 2007.

\bibitem[\protect\citeauthoryear{Lattimore and
  Szepesvári}{2020}]{lattimore2020bandit}
Tor Lattimore and Csaba Szepesvári.
\newblock {\em Bandit Algorithms}.
\newblock Cambridge University Press, 2020.

\bibitem[\protect\citeauthoryear{{Liang} and {Qu}}{2013}]{liang2013portfolio}
J.~J. {Liang} and B.~Y. {Qu}.
\newblock Large-scale portfolio optimization using multiobjective dynamic
  mutli-swarm particle swarm optimizer.
\newblock In {\em 2013 IEEE Symposium on Swarm Intelligence}, 2013.

\bibitem[\protect\citeauthoryear{Mas-Colell \bgroup \em et al.\egroup
  }{1995}]{mascolell:book}
Andreu Mas-Colell, Micheal~D. Whinston, and Jerry~R. Green.
\newblock {\em Microeconomic Theory}.
\newblock Oxford University Press, 1995.

\bibitem[\protect\citeauthoryear{{McFadden}}{1974}]{mcfadden_condlogit:1974}
Daniel {McFadden}.
\newblock Conditional logit analysis of qualitative choice behavior.
\newblock In {\em Frontiers in Econometrics}. 1974.

\bibitem[\protect\citeauthoryear{Paria \bgroup \em et al.\egroup
  }{2019}]{paria19flexible}
Biswajit Paria, Kirthevasan Kandasamy, and Barnab{\'{a}}s P{\'{o}}czos.
\newblock A flexible framework for multi-objective bayesian optimization using
  random scalarizations.
\newblock In {\em UAI}, 2019.

\bibitem[\protect\citeauthoryear{Roijers \bgroup \em et al.\egroup
  }{2017}]{roijers17interactive}
Diederik~M. Roijers, Luisa~M. Zintgraf, and Ann Now{\'{e}}.
\newblock Interactive thompson sampling for multi-objective multi-armed
  bandits.
\newblock In {\em ADT}, 2017.

\bibitem[\protect\citeauthoryear{Rosenbaum and
  Rubin}{1983}]{rosenbaum1983central}
Paul~R. Rosenbaum and Donald~B. Rubin.
\newblock {The central role of the propensity score in observational studies
  for causal effects}.
\newblock {\em Biometrika}, 1983.

\bibitem[\protect\citeauthoryear{Strehl \bgroup \em et al.\egroup
  }{2010}]{strehl2010learning}
Alex Strehl, John Langford, Lihong Li, and Sham~M Kakade.
\newblock Learning from logged implicit exploration data.
\newblock In {\em NeurIPS}, 2010.

\bibitem[\protect\citeauthoryear{Swaminathan and
  Joachims}{2015}]{adith2015counterfactual}
Adith Swaminathan and Thorsten Joachims.
\newblock Counterfactual risk minimization: Learning from logged bandit
  feedback.
\newblock {\em ICML}, 2015.

\bibitem[\protect\citeauthoryear{Tversky and Kahneman}{1974}]{tver-kahn:1974}
Amos Tversky and Daniel Kahneman.
\newblock Judgment under uncertainty: Heuristics and biases.
\newblock {\em Science}, 1974.

\bibitem[\protect\citeauthoryear{Viappiani and
  Boutilier}{2010}]{viappiani:nips2010}
Paolo Viappiani and Craig Boutilier.
\newblock Optimal {Bayesian} recommendation sets and myopically optimal choice
  query sets.
\newblock In {\em NeurIPS}, 2010.

\bibitem[\protect\citeauthoryear{Wong}{1994}]{wong1994comparing}
Weng~Kee Wong.
\newblock Comparing robust properties of a, d, e and g-optimal designs.
\newblock {\em Computational Statistics and Data Analysis}, 1994.

\bibitem[\protect\citeauthoryear{Yahyaa and Manderick}{2015}]{yahyaa15thompson}
Saba~Q. Yahyaa and Bernard Manderick.
\newblock Thompson sampling for multi-objective multi-armed bandits problem.
\newblock In {\em ESANN}, 2015.

\bibitem[\protect\citeauthoryear{Zhang and Golovin}{2020}]{zhang20random}
Richard Zhang and Daniel Golovin.
\newblock Random hypervolume scalarizations for provable multi-objective black
  box optimization.
\newblock In {\em ICML}, 2020.

\bibitem[\protect\citeauthoryear{Zitzler \bgroup \em et al.\egroup
  }{2000}]{zitzler2000comparison}
Eckart Zitzler, Kalyanmoy Deb, and Lothar Thiele.
\newblock Comparison of multiobjective evolutionary algorithms: Empirical
  results.
\newblock {\em Evol. Comput.}, 2000.

\end{thebibliography}
